\documentclass{article}

\usepackage{amsmath}
\usepackage{amsfonts}
\usepackage{bm}
\usepackage{authblk}
\usepackage{graphicx}
\usepackage{url}
\usepackage{hyperref}
\usepackage{listings}
\usepackage{xcolor}

\usepackage{times}
\usepackage{helvet}
\usepackage{courier}
\usepackage{subfig}
\usepackage{float}
\usepackage{algorithm, algorithmic}
\usepackage{amssymb}
\usepackage{amsthm}
\usepackage{wrapfig}
\usepackage{multirow}

\definecolor{mygreen}{rgb}{0,0.6,0}
\definecolor{mygray}{rgb}{0.5,0.5,0.5}
\definecolor{mymauve}{rgb}{0.58,0,0.82}

\begin{document}

\title{Unsupervised Multi-modal\\Neural Machine Translation}

\author[a]{Yuanhang Su$\thanks{indicates equal contribution. Work performed while Yuanhang Su was an internship at Alibaba.}$}
\author[b]{Kai Fan$^*$\thanks{Corresponding author: \url{k.fan@alibaba-inc.com}.}}
\author[b]{Nguyen Bach}
\author[a]{C.-C. Jay Kuo}
\author[b]{Fei Huang}

\affil[a]{University of South California}
\affil[b]{Alibaba Group (U.S.) Inc.}

\maketitle

\begin{abstract}
Unsupervised neural machine translation (UNMT) has recently achieved remarkable results \cite{lample2018phrase} with only large monolingual corpora in each language. 
However, the uncertainty of associating target with source sentences makes UNMT theoretically an ill-posed problem. 
This work investigates the possibility of utilizing images for disambiguation to improve the performance of UNMT. 
Our assumption is intuitively based on the invariant property of image, i.e., the description of the same visual content by different languages should be approximately similar. 
We propose an unsupervised multi-modal machine translation (UMNMT) framework based on the language translation cycle consistency loss conditional on the image, targeting to learn the bidirectional multi-modal translation simultaneously. 
Through an alternate training between multi-modal and uni-modal, our inference model can translate with or without the image. 
On the widely used Multi30K dataset, the experimental results of our approach are significantly better than those of the text-only UNMT on the 2016 test dataset. 
\end{abstract}

\section{Introduction}

Our long-term goal is to build intelligent systems that can perceive their visual environment and understand the linguistic information, and further make an accurate translation inference to another language. 
Since image has become an important source for humans to learn and acquire knowledge (e.g. video lectures,  \cite{alayrac2016unsupervised,kuehne2014language,zhu2018generative}), the visual signal might be able to disambiguate certain semantics. 
One way to make image content easier and faster to be understood by humans is to combine it with narrative description that can be self-explainable. 
This is particularly important for many natural language processing (NLP) tasks as well, such as image caption \cite{vinyals2015show} and some task-specific translation--sign language translation \cite{camgoz2018neural}.
However, \cite{specia2016shared} demonstrates that most multi-modal translation algorithms are not significantly better than an off-the-shelf text-only machine translation (MT) model for the Multi30K dataset \cite{m30k}.
There remains an open question about how translation models should take advantage of visual context, because from the perspective of information theory, the mutual information of two random variables $I(X,Y)$ will always be no greater than $I(X;Y,Z)$, due to the following fact.
\begin{align}
 & I(X;Y,Z) - I(X;Y) \nonumber \\
=& KL(p(X,Y,Z)\|p(X|Y)p(Z|Y)p(Y)) \label{eq:mutual} 
\end{align}
where the Kullback-Leibler (KL) divergence is non-negative. 
This conclusion makes us believe that the visual content will hopefully help the translation systems.

\begin{figure}[t]
\includegraphics[width=\textwidth]{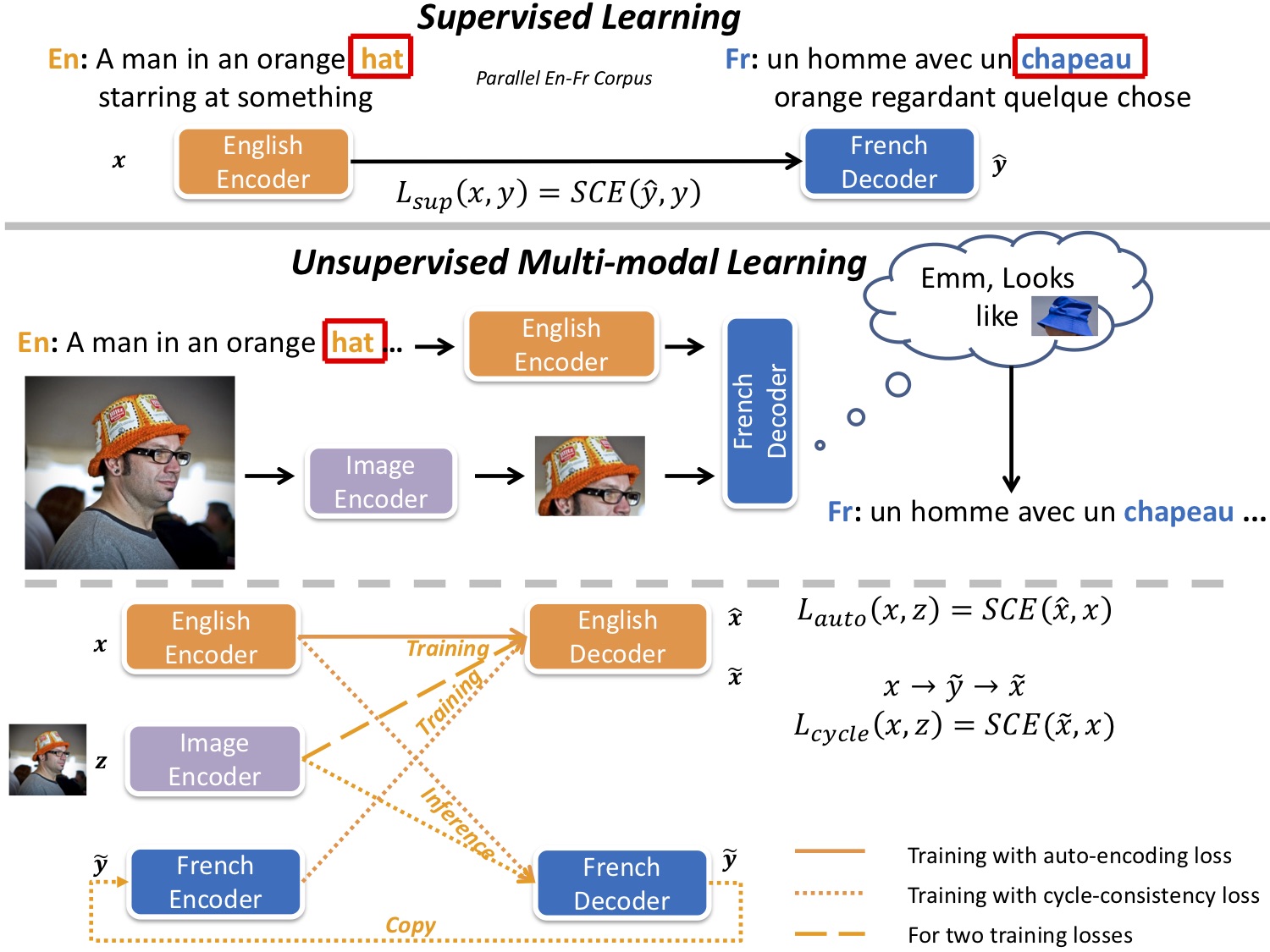}
\caption{Illustration of our proposed approach. We leverage the designed loss function to tackle a supervised task with the unsupervised dataset only. SCE means sequential cross-entropy.}
\label{fig:intro}
\end{figure}

Since the standard paradigm of multi-modal translation always considers the problem as a supervised learning task, the parallel corpus is usually sufficient to train a good translation model, and the gain from the extra image input is very limited. 
Moreover, the scarcity of the well formed dataset including both images and the corresponding multilingual text descriptions is also another constraint to prevent the development of more scaled models. 
In order to address this issue, we propose to formulate the multi-modal translation problem as an unsupervised learning task, which is closer to real applications. 
This is particularly important given the massive amounts of paired image and text data being produced everyday (e.g., news title and its illustrating picture).

Our idea is originally inspired by the text-only unsupervised MT (UMT) \cite{conneau2017word,lample2017unsupervised,lample2018phrase}, investigating whether it is possible to train a general MT system without any form of supervision. 
As \cite{lample2018phrase} discussed, the text-only UMT is fundamentally an ill-posed problem, since there are potentially many ways to associate target with source sentences. 
Intuitively, since the visual content and language are closely related, the image can play the role of a pivot ``language" to bridge the two languages without paralleled corpus, making the problem ``more well-defined" by reducing the problem to supervised learning. 
However, unlike the text translation involving word generation (usually a discrete distribution), the task to generate a dense image from a sentence description itself is a challenging problem \cite{mansimov2015generating}. 
High quality image generation usually depends on a complicated or large scale neural network architecture \cite{reed2016generative,fan2017inversenet,Tao18attngan}. 
Thus, it is not recommended to utilize the image dataset as a pivot ``language" \cite{chen2018zero}. 
Motivated by the cycle-consistency \cite{CycleGAN2017}, we tackle the unsupervised translation with a multi-modal framework which includes two sequence-to-sequence encoder-decoder models and one shared image feature extractor. 
We don't introduce the adversarial learning via a discriminator because of the non-differentiable $\arg\max$ operation during word generation. 
With five modules in our framework, there are multiple data streaming paths in the computation graph, inducing the auto-encoding loss and cycle-consistency loss, in order to achieve the unsupervised translation.

Another challenge of unsupervised multi-modal translation, and more broadly for general multi-modal translation tasks, is the need to develop a reasonable multi-source encoder-decoder model that is capable of handling multi-modal documents. 
Moreover, during training and inference stages, it is better to process the mixed data format including both uni-modal and multi-modal corpora.

First, this challenge highly depends on the attention mechanism across different domains. 
Recurrent Neural Networks (RNN) and Convolutional Neural Networks (CNN) are naturally suitable to encode the language text and visual image respectively; however, encoded features of RNN has autoregressive property which is different from the local dependency of CNN. 
The multi-head self-attention transformer \cite{vaswani2017attention} can mimic the convolution operation, and allow each head to use different linear transformations, where in turn different heads can learn different relationships. 
Unlike RNN, it reduces the length of the paths of states from the higher layer to all states in the lower layer to one, and thus facilitates more effective learning.
For example, the BERT model \cite{devlin2018bert}, that is completely built upon self-attention, has achieved remarkable performance in 11 natural language tasks.
Therefore, we employ transformer in both the text encoder and decoder of our model, and design a novel joint attention mechanism to simulate the relationships among the three domains.
Besides, the mixed data format requires the desired attention to support the flexible data stream. 
In other words, the batch fetched at each iteration can be either uni-modal text data or multi-modal text-image paired data, allowing the model to be adaptive to various data during inference as well.

Succinctly, our contributions are three-fold: 

\textbf{(1)} We formuate the multi-modal MT problem as unsupervised setting that fits the real scenario better and propose an end-to-end transformer based multi-modal model.  

\textbf{(2)} We present two technical contributions: successfully train the proposed model with auto-encoding and cycle-consistency losses, and design a controllable attention module to deal with both uni-modal and multi-modal data.

\textbf{(3)} We apply our approach to the Multilingual Multi30K dataset in English$\leftrightarrow$French and English$\leftrightarrow$German translation tasks, and the translation output and the attention visualization show the gain from the extra image is significant in the unsupervised setting.

\section{Related Work}

We place our work in context by arranging several prior popular topics, along the the axes of UMT, image caption and multi-modal MT.

\textbf{Unsupervised Machine Translation} Existing methods in this area  \cite{artetxe2017unsupervised,lample2017unsupervised,lample2018phrase} are mainly modifications of encoder-decoder schema. 
Their key ideas are to build a common latent space between the two languages (or domains) and to learn to translate by reconstructing in both domains. 
The difficulty in multi-modal translation is the involvement of another visual domain, which is quite different from the language domain. 
The interaction between image and text are usually not symmetric as two text domains. 
This is the reason why we take care of the attention module cautiously.

\textbf{Image Caption} Most standard image caption models are built on CNN-RNN based encoder-decoder framework \cite{karpathy2015deep,vinyals2015show}, where the visual features are extracted from CNN and then fed into RNN to output word sequences as captions.
Since our corpora contain image-text paired data, our method also draws inspiration from image caption modeling. 
Thus, we also embed the image-caption model within our computational graph, whereas the transformer architecture is adopted as a substitution for RNN.

\textbf{Multi-modal Machine Translation} This problem is first proposed by \cite{specia2016shared} on the WMT16 shared task at the intersection of natural language processing and computer vision. 
It can be considered as building a multi-source encoder on top of either MT or image caption model, depending on the definition of extra source. 
Most Multi-modal MT research still focuses on the supervised setting, while \cite{chen2018zero,nakayama2017zero}, to our best knowledge, are the two pioneering works that consider generalizing the Multi-modal MT to an unsupervised setting. 
However, their setup puts restrictions on the input data format. 
For example, \cite{chen2018zero} requires the training data to be image text pair but the inference data is text-only input, and \cite{nakayama2017zero} requires image text pair format for both training and testing. 
These limit the model scale and generalization ability, since large amount of monolingual corpora is more available and less expensive. 
Thus, in our model, we specifically address this issue with controllable attention and alternative training scheme.

\section{Methodology}
In this section we first briefly describe the main MT systems that our method is built upon and then elaborate on our approach.

\subsection{Neural Machine Translation}
If a bilingual corpus is available, given a source sentence $\mathbf{x}=(x_1,...,x_n)$ of $n$ tokens, and a translated target sentence $\mathbf{y}=(y_1,...,y_m)$ of $m$ tokens, where $(\mathbf{x}, \mathbf{y})\in\mathcal{X}\times\mathcal{Y}$, the NMT model aims at maximizing the likelihood,
\begin{equation}\label{eq:nmt_likelihood}
p(\mathbf{y}|\mathbf{x}) = \sum_{t=1}^m p(y_t|\mathbf{y}_{<t}, \mathbf{x}) .
\end{equation}
The attention based sequence-to-sequence encoder-decoder architecture \cite{bahdanau2014neural,wu2016google,gehring2017convolutional,vaswani2017attention} is usually employed to parameterize the above conditional probability. 

The encoder reads the source sentence and outputs the hidden representation vectors for each token, $\{\mathbf{h}_1^e,...,\mathbf{h}_n^e\} = \text{Enc}_x(\mathbf{x})$. 
The attention based decoder is defined in a recurrent way. 
Given the decoder has the summarized representation vector $\mathbf{h}_{t}^d=\text{Dec}_y(\mathbf{y}_{<t}, \mathbf{x})$ at time stamp $t$, the model produces a context vector $\mathbf{c}_t=\sum_{j=1}^n \alpha_i\mathbf{h}_j^e$ based on an alignment model, $\{\alpha_1,...,\alpha_n\} = \text{Align}(\mathbf{h}_{t}^d, \{\mathbf{h}_1^e,...,\mathbf{h}_n^e\})$, such that $\sum_{j=1}^n\alpha_j=1$. 
Therefore, the conditional probability to predict the next token can be written as,
\begin{equation}\label{eq:cond_prob}
p(y_t|\mathbf{y}_{<t}, \mathbf{x}) = \text{softmax}(g(\mathbf{c}_t, y_{t-1}, \mathbf{h}_{t-1}^d)) .
\end{equation}
in which $g(\cdot)$ denotes a non-linear function extracting features to predict the target. 
The encoder and decoder model described here is in a general formulation, not constrained to be RNN \cite{bahdanau2014neural} or transformer architecture \cite{vaswani2017attention}.

\begin{figure*}[t]
\includegraphics[width=\textwidth]{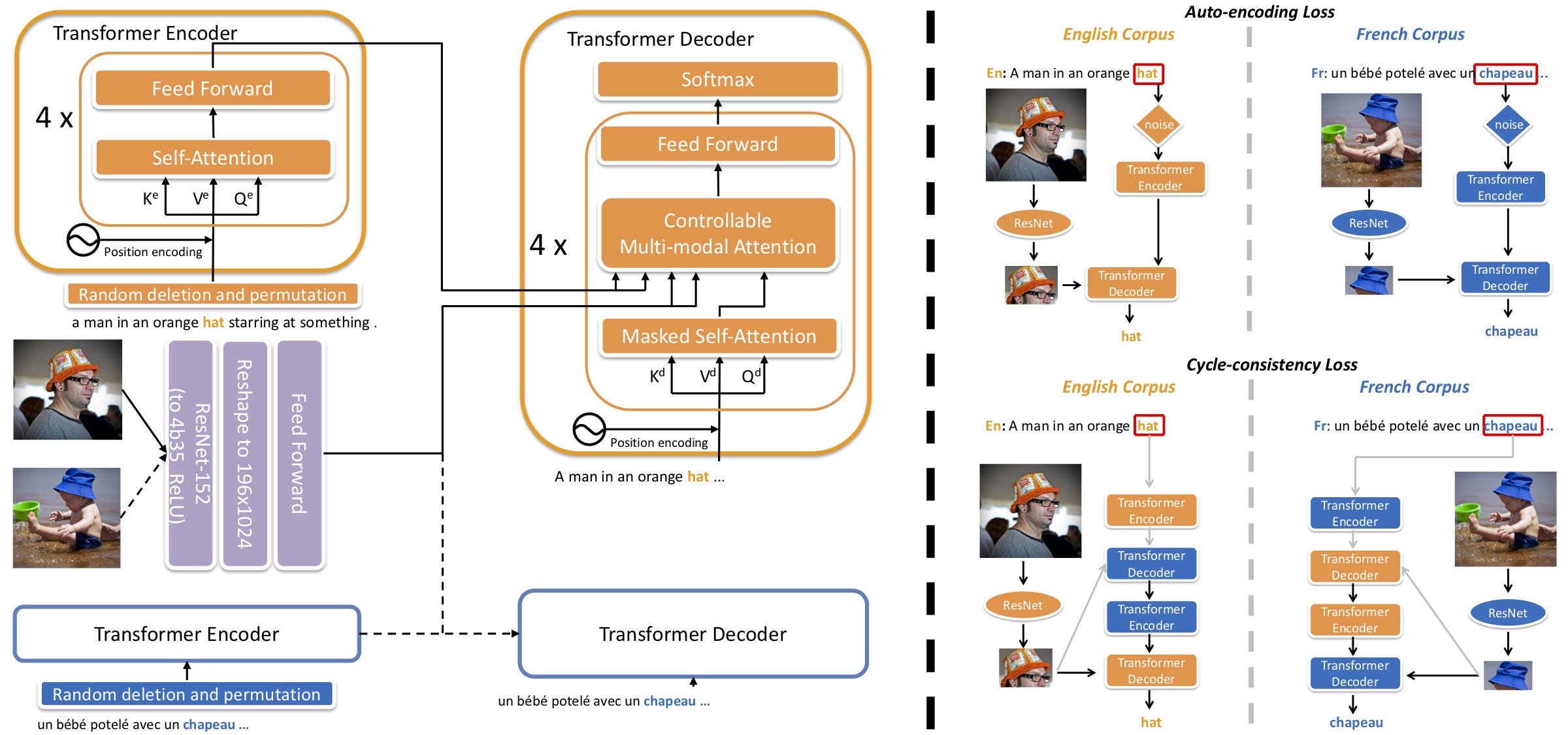}
\caption{Model overview. Left Panel: The detailed unsupervised multi-modal neural machine translation model includes five modules, two transformer encoder, two transformer decoder and one ResNet encoder. Some detailed network structures within the transformer, like skip-connection and layer normalization, are omitted for clarity. Right Panel: The entire framework consists of four training paths: the gray arrows in the paths for cycle-consistency loss indicate the model is under inference mode. \textit{E.g.}, the time step decoding for token ``hat" is illustrated.}
\label{fig:model}
\end{figure*}

\subsection{Multi-modal Neural Machine Translation}
In this task, an image $\mathbf{z}$ and the description of the image in two different languages form a triplet $(\mathbf{x}, \mathbf{y}, \mathbf{z})\in\mathcal{X}\times\mathcal{Y}\times\mathcal{I}$. 
Thus, the problem naturally becomes maximizing the new likelihood $p(\mathbf{y}|\mathbf{x},\mathbf{z})$. 
Though the overall framework of such a translation task is still the encoder-decoder architecture, the detailed feature extractor and attention module can vary greatly, due to the extra source image. 

The traditional approach \cite{specia2016shared,elliott2017findings} is to encode the source text and the image separately and combine them at the high level features, where the image feature map can be represented as $\{\mathbf{h}_{1}^i,...,\mathbf{h}_k^i\}=\text{Enc}_z(\mathbf{z})$ and $\text{Enc}_z$ is usually a truncated image classification model, such as Resnet \cite{he2016deep}. 
Notice that unlike the number of the text features is exactly the number of tokens in the source, the number of the image features $k$ depends on the last layer in the truncated network. 
Then, the context vector is computed via an attention model, 
\begin{equation}\label{eq:mmt_context}
\mathbf{c}_t=\text{Attention}(\mathbf{h}_{t}^d, \{\mathbf{h}_1^e,...,\mathbf{h}_n^e\}, \{\mathbf{h}_{1}^i,...,\mathbf{h}_k^i\})
\end{equation}
Since three sets of features appear in Eq~(\ref{eq:mmt_context}), there are more options of the attention mechanism than text-only NMT. 
The decoder can remain the same in the recurrent fashion.

\subsection{Unsupervised Learning}
The unsupervised problem requires a new problem definition. 
On both the source and the target sides, only monolingual documents are presented in the training data, i.e., the data comes in the paired form of $(\mathbf{x}, \mathbf{z})\in\mathcal{X}\times\mathcal{I}$ and $(\mathbf{y}, \mathbf{z})\in\mathcal{Y}\times\mathcal{I}$. 
The triplet data format is no longer available. 
The purpose is to learn a multi-modal translation model $\mathcal{X}\times\mathcal{I} \rightarrow \mathcal{Y}$ or a text-only one $\mathcal{X} \rightarrow \mathcal{Y}$. 
Note there is no explicit paired information cross two languages, making it impossible to straightforwardly optimize the supervised likelihood.
Fortunately, motivated by the CycleGAN \cite{CycleGAN2017} and the dual learning in \cite{he2016dual}, we can actually learn the translation model for both directions between the source and the target in an unsupervised way. 
Additionally, we can even make the multi-modal and uni-modal inference compatible with deliberate fine-tuning strategy. 

\subsection{Auto-Encoding Loss}
As Figure~\ref{fig:model} illustrates, there are five main modules in the overall architecture, two encoders and two decoders for the source and target languages, and one extra image encoder. 
Since the lack of triplet data, we can only build the first two following denoised auto-encoding losses without involving the paired $\mathbf{x}$ and $\mathbf{y}$,
\begin{align}
\mathcal{L}_{\text{auto}}(\mathbf{x}, \mathbf{z}) = SCE(\text{Dec}_{x}(\text{Enc}_{x}(\mathbf{x}), \text{Enc}_{z}(\mathbf{z})), \mathbf{x}) \label{eq:auto1} \\
\mathcal{L}_{\text{auto}}(\mathbf{y}, \mathbf{z}) = SCE(\text{Dec}_{y}(\text{Enc}_{y}(\mathbf{y}), \text{Enc}_{z}(\mathbf{z})), \mathbf{x}) \label{eq:auto2} 
\end{align}
where $SCE(\cdot,\cdot)$ represents sequential cross-entropy loss. 
We use ``denoised" loss here, because the exact auto-encoding structure will likely force the language model learning a word-to-word copy network. 
The image is seemingly redundant since the text input contains the entire information for recovery. 
However, it is not guaranteed that our encoder is lossless, so the image is provided as an additional supplement to reduce the information loss. 

\subsection{Cycle-Consistency Loss}
The auto-encoding loss can, in theory, learn two functional mappings $\mathcal{X}\times\mathcal{I}\rightarrow\mathcal{X}$ and $\mathcal{Y}\times\mathcal{I}\rightarrow\mathcal{Y}$ via the supplied training dataset. 
However, the two mappings are essentially not our desiderata, even though we can switch the two decoders to build our expected mappings, e.g., $\mathcal{X}\times\mathcal{I}\rightarrow\mathcal{Y}$. 
The crucial problem is that the transferred mappings achieved after switching decoders lack supervised training, since no regularization pushes the latent encoding spaces aligned between the source and target.

We argue that this issue can be tackled by another two cycle-consistency properties (note that we use the square brackets $[]$ below to denote the inference mode, meaning no gradient back-propagation through such operations), 
\begin{align}
\text{Dec}_{x}(\text{Enc}_{y}(\text{Dec}_{y}[\text{Enc}_{x}(\mathbf{x}), \text{Enc}_{z}(\mathbf{z})]), \text{Enc}_{z}(\mathbf{z})) \approx \mathbf{x} \label{eq:cycle1} \\
\text{Dec}_{y}(\text{Enc}_{x}(\text{Dec}_{x}[\text{Enc}_{y}(\mathbf{y}), \text{Enc}_{z}(\mathbf{z})]), \text{Enc}_{z}(\mathbf{z})) \approx \mathbf{y} \label{eq:cycle2}
\end{align}
The above two properties seem complicated, but we will decompose them step-by-step to see its intuition, which are also the key to make the auto-encoders translation models across different languages. 
Without loss of generality, we use Property (\ref{eq:cycle1}) as our illustration, where the same idea is applied to (\ref{eq:cycle2}). 
After encoding the information from source and image as the high level features, the encoded features are fed into the decoder of another language (i.e. target language), thus obtaining an inferred target sentence,
\begin{equation}\label{eq:inferred_y}
\tilde{\mathbf{y}} = F_{xz\rightarrow y}(\mathbf{x},\mathbf{z}) \triangleq \text{Dec}_{y}[\text{Enc}_{x}(\mathbf{x}), \text{Enc}_{z}(\mathbf{z})] .
\end{equation}
Unfortunately, the ground truth $\mathbf{y}$ corresponding to the input $\mathbf{x}$ or $\mathbf{z}$ is unknown, so we cannot train $F_{xz\rightarrow y}$ at this time. 
However, since $\mathbf{x}$ is the golden reference, we can construct the pseudo supervised triplet $(\mathbf{x}, \tilde{\mathbf{y}}, \mathbf{z})$ as the augmented data to train the following model,
\begin{equation}\label{eq:predict_x}
F_{yz\rightarrow x}(\tilde{\mathbf{y}},\mathbf{z}) \triangleq \text{Dec}_{x}(\text{Enc}_{y}(\tilde{\mathbf{y}}), \text{Enc}_{z}(\mathbf{z})).
\end{equation}
Note that the pseudo input $\tilde{\mathbf{y}}$ can be considered as the corrupted version of the unknown $\mathbf{y}$. 
The noisy training step makes sense because injecting noise to the input data is a common trick to improve the robustness of model even for traditional supervised learning \cite{srivastava2014dropout,wang2018switchout}.
Therefore, we incentivize this behavior using the cycle-consistency loss,
\begin{align}\label{eq:cycle_loss}
\mathcal{L}_{\text{cyc}}(\mathbf{x},\mathbf{z}) = SCE(F_{yz\rightarrow x}(F_{xz\rightarrow y}(\mathbf{x},\mathbf{z}),\mathbf{z}), \mathbf{x}) .
\end{align}
This loss indicates the cycle-consistency $(\ref{eq:cycle1})$, and the mapping $\mathcal{Y}\times\mathcal{I}\rightarrow\mathcal{X}$ can be successfully refined.

\subsection{Controllable Attention}
In additional to the loss function, another important interaction between the text and image domain should focus on the decoder attention module. 
In general, we proposal to extend the traditional encoder-decoder attention to a multi-domain attention.
\begin{align}\label{eq:tri_att}
\mathbf{c}_t = \text{Att}(\mathbf{h}_t^d, \mathbf{h}^e) + \lambda_1\text{Att}(\mathbf{h}_t^d, \mathbf{h}^i) + \lambda_2 \text{Att}(\mathbf{h}_t^d, \mathbf{h}^e, \mathbf{h}^i)
\end{align}
where $\lambda_1$ and $\lambda_2$ can be either 1 or 0 during training, depending on whether the fetched batch includes image data or not. 
For example, we can easily set up a flexible training scheme by alternatively feeding the monolingual language data and text-image multimodal data to the model. 
A nice byproduct of this setup allows us to successfully make a versatile inference with or without image, being more applicable to real scenarios. 

In practice, we utilize the recent developed self-attention mechanism \cite{vaswani2017attention} as our basic block, the hidden states contain three sets of vectors $Q, K, V$, representing queries, keys and values. 
Therefore, our proposed context vector can be rewritten as,
\begin{align}
\mathbf{c}_t &= \text{softmax}\left(\frac{Q_t^d (K^e)^\top}{\sqrt{d}}\right)V^e + \lambda_1 \text{softmax}\left(\frac{Q_t^d (K^i)^\top}{\sqrt{d}}\right)V^i \nonumber \\
             &+ \lambda_2 \text{softmax}\left(\frac{Q_t^d (K^{ei})^\top}{\sqrt{d}}\right)V^{ei} \nonumber \\ 
             &+ \lambda_2 \text{softmax}\left(\frac{Q_t^d (K^{ie})^\top}{\sqrt{d}}\right)V^{ie}
\end{align}
where $d$ is the dimensionality of keys, and $[K^{ei}, V^{ei}]=\text{FFN}\left(\text{softmax}\left(\frac{Q^e(K^i)^\top}{\sqrt{d}}\right)V^i \right)$ means the attention from text input to image input, and $[K^{ie}, V^{ie}]$ represents the symmetric attention in the reverse direction. 
Note the notation $Q^e$ has no subscript and denotes as a matrix, indicating the softmax is row-wise operation. 
In practice, especially for Multi30K dataset, we found $\lambda_2$ is less important and $\lambda_2=0$ brings no harm to the performance. 
Thus, we always set it as 0 in our experiments, but non-zero $\lambda_2$ may be helpful in other cases.

\section{Experiments}\label{sec:exp}

\subsection{Training and Testing on Multi30K}\label{sec:exp_setup}
We evaluate our model on Multi30K \cite{m30k} 2016 test set of English$\leftrightarrow$French (En$\leftrightarrow$Fr) and English$\leftrightarrow$German (En$\leftrightarrow$De) language pairs. 
This dataset is a multilingual image caption dataset with 29000 training samples of images and their annotations in English, German, French \cite{m30k_fr} and Czech \cite{m30k_cz}. 
The validation set and test set have 1014 and 1000 samples respectively. 
To ensure the model never sees any paired sentences information (which is an unlikely scenario in practice), we randomly split half of the training and validation sets for one language and use the complementary half for the other. 
The resulting corpora is denoted as \textit{M30k-half} with 14500 and 507 training and validation samples respectively.

To find whether the image as additional information used in the training and/or testing stage can bring consistent performance improvement, we train our model in two different ways, each one has train with text only (-txt) and train with text+image (-txt-img) modes. 
We would compare the best performing training method to the state-of-the-art, and then do side-by-side comparison between them:

{\bf Pre-large (P)}: To leverage the controllable attention mechanism for exploring the linguistic information in the large monolingual corpora, we create text only pre-training set by combining the first 10 million sentences of the WMT News Crawl datasets from 2007 to 2017 with 10 times M30k-half. This ends up in a large text only dataset of 10145000 unparalleled sentences in each language. 
{\bf P-txt}: We would then pre-train our model without the image encoder on this dataset and use the M30k-half validation set for validation. {\bf P-txt-img}: Once the text-only model is pre-trained, we then use it for the following fine-tuning stage on M30k-half. 
Except for the image encoder, we initialize our model with the pre-trained model parameters.
The image encoder uses pre-trained ResNet-152 \cite{he2016deep}. 
The error gradient does not back-propagate to the original ResNet network.

{\bf Scratch (S)}: We are also curious about the role of image can play when no pre-training is involved. 
We train from scratch using text only ({\bf S-txt}) and text with corresponding image ({\bf S-txt-img}) on M30k-half.

\subsection{Implementation Details and Baseline Models}
The text encoder and decoder are both 4 layers transformers with dimensionality 512, and for the related language pair, we share the first 3 layers of transformer for both encoder and decoder. 
The image encoder is the truncated ResNet-152 with output layer res4b35\_relu, and the parameters of ResNet are freezing during model optimization. 
Particularly, the feature map $14\times14\times1024$ of layer res4b35\_relu is flattened to $196\times1024$ so that its dimension is consistent with the sequential text encoder output.
The actual losses (\ref{eq:auto1}) and (\ref{eq:auto2}) favor a standard denoising auto-encoders: the text input is perturbed with deletion and local permutation; the image input is corrupted via dropout. 
We use the same word preprocessing techniques (Moses tokenization, BPE, binarization, fasttext word embedding on training corpora, etc.) as reported in \cite{lample2018phrase}, please refer to the relevant readings for further details.

We would like to compare the proposed UMNMT model to the following UMT models. 
\begin{itemize}
\setlength{\itemsep}{0pt}
	\item MUSE \cite{conneau2017word}: It is an unsupervised word-to-word translation model. The embedding matrix is trained on large scale wiki corpora.
	\item Game-NMT \cite{chen2018zero}: It is a multi-modal zero-source UMT method trained using reinforcement learning.
	\item UNMT-text \cite{lample2017unsupervised}: It is a mono-modal UMT model which only utilize text data and it is pretrained on synthetic paired data generated by MUSE.
\end{itemize}

\begin{table}[t]
\small
\centering
\begin{tabular}{l|c|c|c|c}
\hline
Models & En$\rightarrow$  Fr & Fr$\rightarrow$ En & En$\rightarrow$  De & De$\rightarrow$  En  \\ \hline
MUSE & 8.54 & 16.77 & 15.72 & 5.39 \\ 
Game-NMT & - & - & 16.6 & 19.6 \\ 
UNMT-text & 32.76  & 32.07 & 22.74 & 26.26 \\ \hline\hline
S-txt  & 6.01 & 6.75 & 6.27 & 6.81 \\
S-txt-img  & 9.40 & 10.04 & 8.85 & 9.97 \\\hline
P-txt & 37.20 & 38.51 & 20.97 & 25.00 \\
P-txt-img & {\bf 39.79} & {\bf 40.53} & {\bf 23.52} & {\bf 26.39}\\\hline
\end{tabular}
\caption{BLEU benchmarking. The numbers of baseline models are extracted from the corresponding references.}
\label{tab:exp_stoa}
\end{table}
\begin{figure}[t]
\hspace*{-0.1in}
\centering
\subfloat{\label{fig:exp_en-fr} \includegraphics[width 
= 0.5\linewidth]{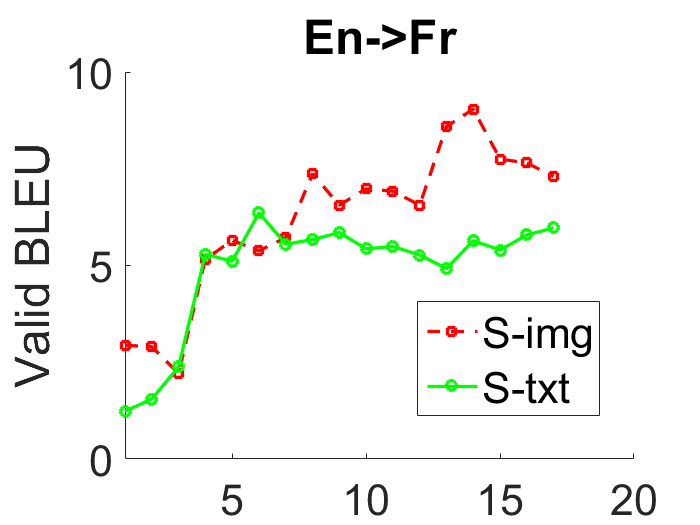}}
\centering
\subfloat{\label{fig:exp_fr-en} \includegraphics[width 
= 0.5\linewidth]{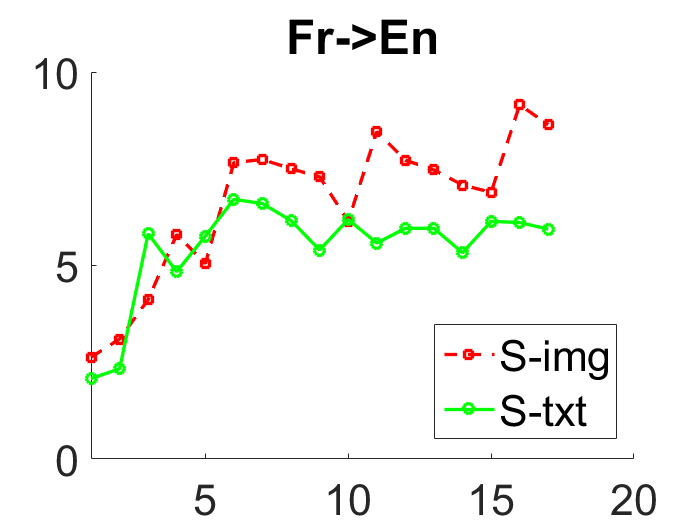}}\\
\centering
\subfloat{\label{fig:exp_en-de} \includegraphics[width 
= 0.5\linewidth]{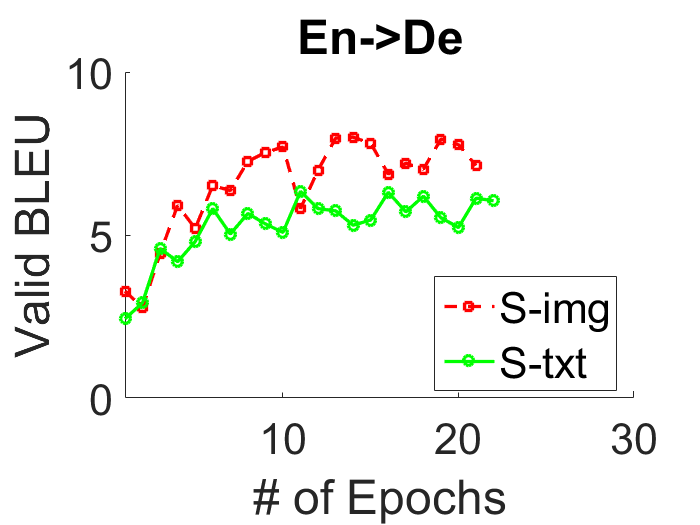}}
\centering
\subfloat{\label{fig:exp_de-en} \includegraphics[width 
= 0.5\linewidth]{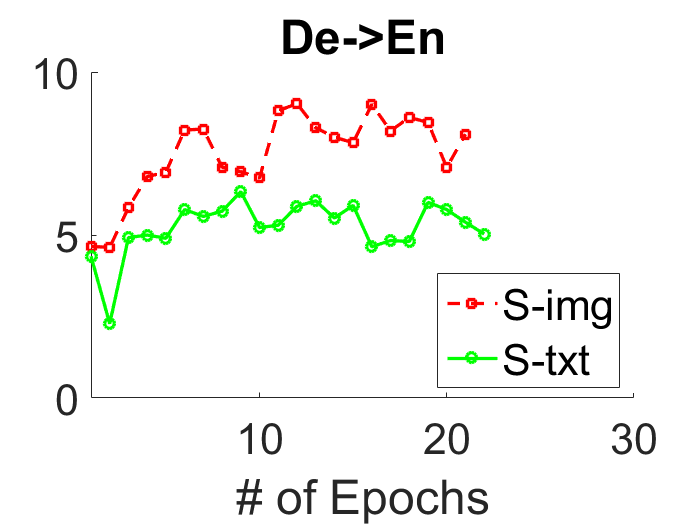}} 
\caption{Validation BLEU comparison between text-only and text+image.} \label{fig:exp_curve}
\end{figure}
\begin{table*}[htbp]
\hspace*{-2.5cm}
\small
\centering
\begin{tabular}{l|c|c|c|c|c|c|c|c|c|c|c|c}
\hline
 & \multicolumn{3}{|c|}{En$\rightarrow$Fr} & \multicolumn{3}{|c|}{Fr$\rightarrow$En} & \multicolumn{3}{|c|}{En$\rightarrow$De} & \multicolumn{3}{|c}{De$\rightarrow$En}\\
 \cline{2-13}
Models & Meteor & Rouge & CIDEr & Meteor & Rouge & CIDEr & Meteor & Rouge & CIDEr & Meteor & Rouge & CIDEr  \\
\hline
\hline
S-txt  & 0.137 & 0.325 & 0.46 & 0.131 & 0.358 & 0.48 & 0.116 & 0.306 & 0.35 & 0.128 & 0.347 & 0.47 \\
S-txt-img  & {\bf 0.149} & {\bf 0.351} & {\bf 0.65} & {\bf 0.155} & {\bf 0.401} & {\bf 0.75} & {\bf 0.138} & {\bf 0.342} & {\bf 0.59} & {\bf 0.156} & {\bf 0.391} & {\bf 0.70} \\\hline
P-txt & 0.337 & 0.652 & 3.36 & 0.364 & 0.689 & 3.41 & 0.254 & 0.539 & 1.99 & 0.284 & 0.585 & 2.20 \\
P-txt-img & {\bf 0.355} & {\bf 0.673} & {\bf 3.65} & {\bf 0.372} & {\bf 0.699} & {\bf 3.61} & {\bf 0.261} & {\bf 0.551} & {\bf 2.13} & {\bf 0.297} & {\bf 0.597} & {\bf 2.36} \\
\hline
\end{tabular}
\caption{UMNMT shows consistent improvement over text-only model across normalized Meteor, Rouge and CIDEr metrics.}
\label{tab:other_metrics}
\end{table*}

\begin{figure}[htbp]
\centering
\includegraphics[width=0.8\textwidth]{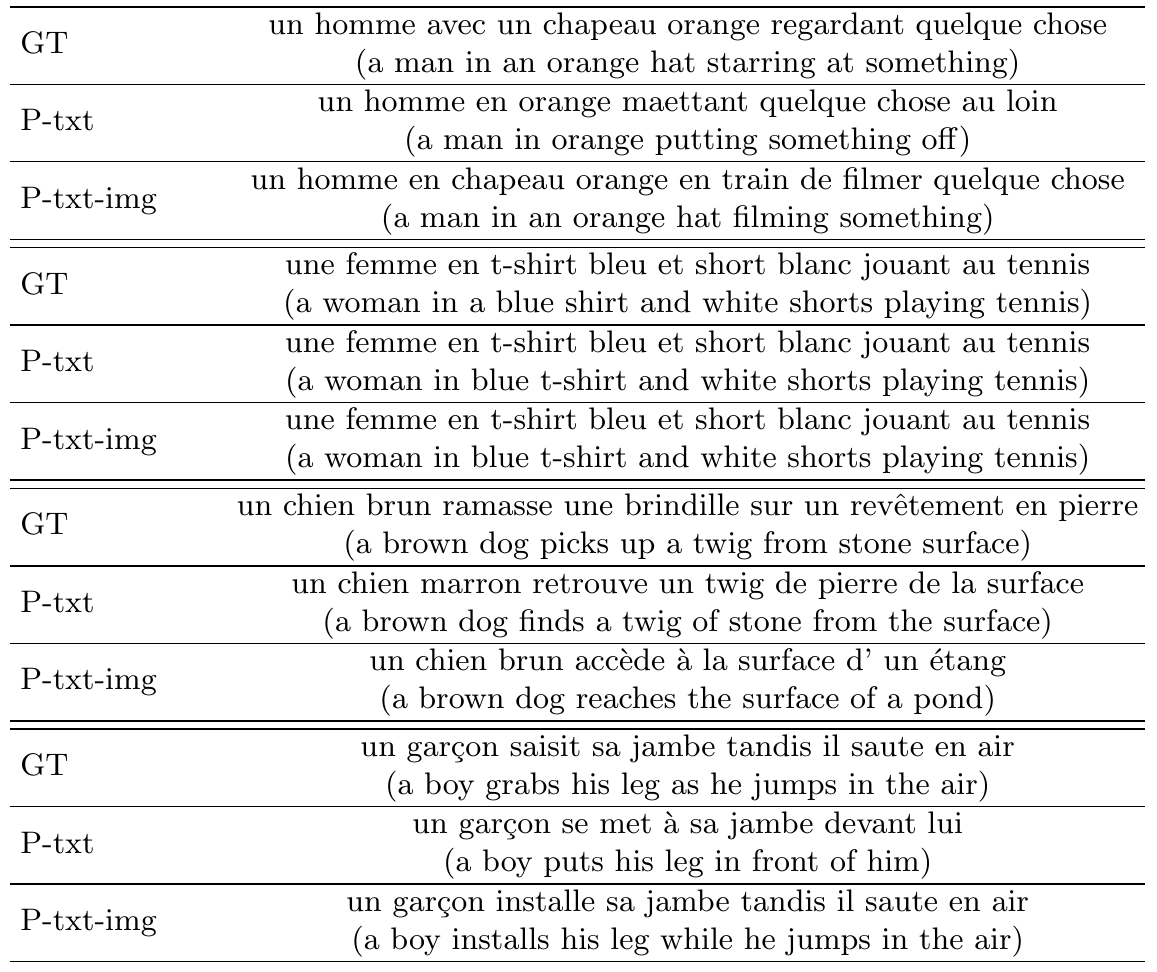}
\caption{Translation results from different models (GT: ground truth)} \label{fig:translations}
\end{figure}

\subsection{Benchmarking with state-of-the-art}\label{sec:exp_res}

In this section, we report the widely used BLEU score of test dataset in Table \ref{tab:exp_stoa} for different MT models. 
Our best model has achieved the state-of-the-art performance by leading more than 6 points in En$\rightarrow$Fr task to the second best. 
Some translation examples are shown in Figure~\ref{fig:translations}. 
There is also close to 1 point improvement in the En$\rightarrow$De task. 
Although pre-training plays a significant role to the final performance, the image also contributes more than 3 points in case of training from scratch (S-txt vs. S-txt-img), and around 2 points in case of fine tuning (P-txt vs. P-txt-img). 
Interestingly, it is observed that the image contributes less performance improvement for pre-training than training from scratch. 
This suggests that there is certain information overlap between the large monolingual corpus and the M30k-half images. 
We also compare the Meteor, Rouge, CIDEr score in Table \ref{tab:other_metrics} and validation BLEU in Figure~\ref{fig:exp_curve} to show the consistent improvement brought by using images.

\begin{figure*}[t]
\centering
\includegraphics[width=0.9\textwidth]{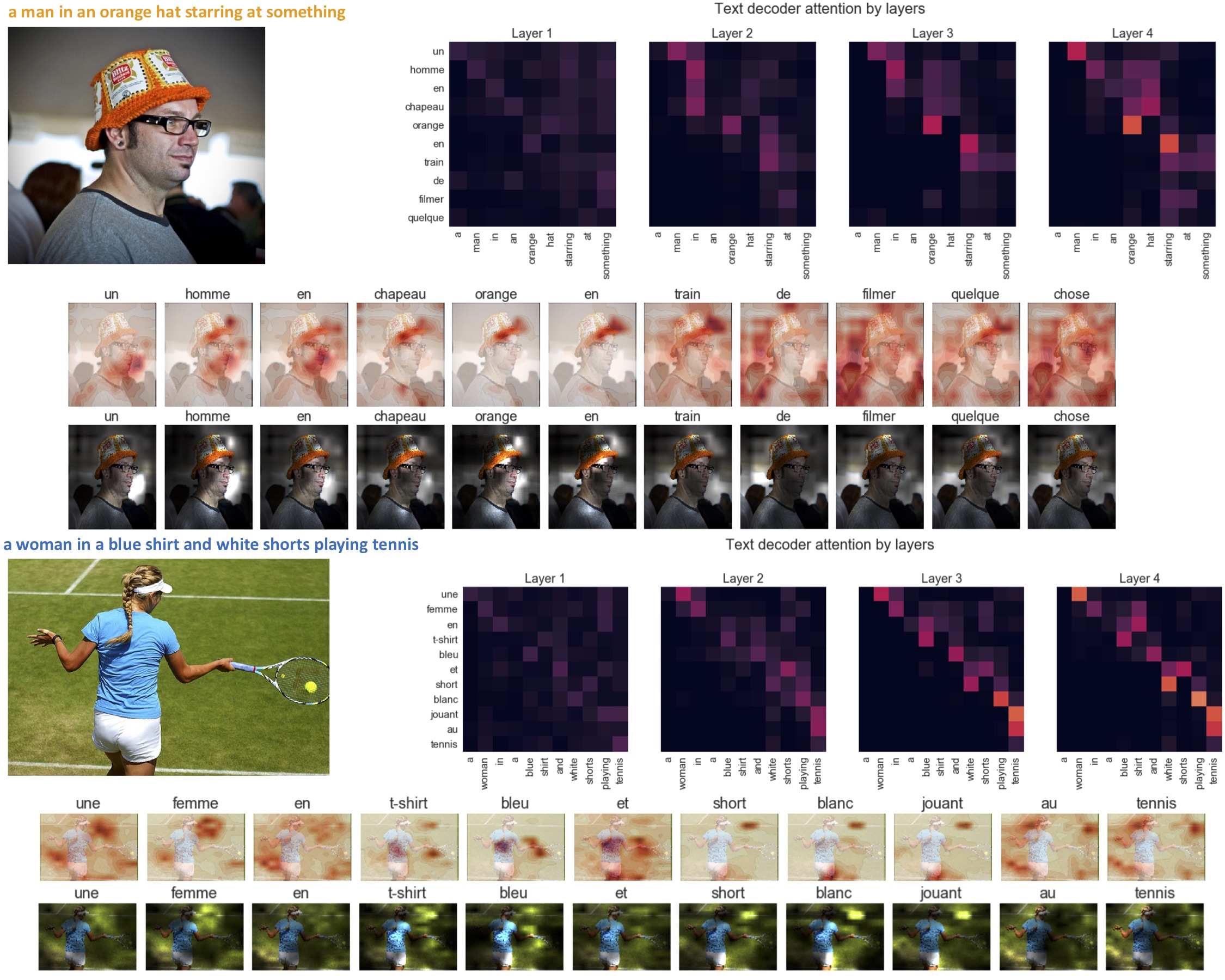}
\caption{Correct attention for \{``humme", ``chapeau", ``orange", ``chose"\} and \{``bleu", ``t-shirt", ``blanc", ``short"\}.} \label{fig:exp_visual_positive}
\end{figure*}

\begin{figure*}[htbp]
\centering
\includegraphics[width=0.88\textwidth]{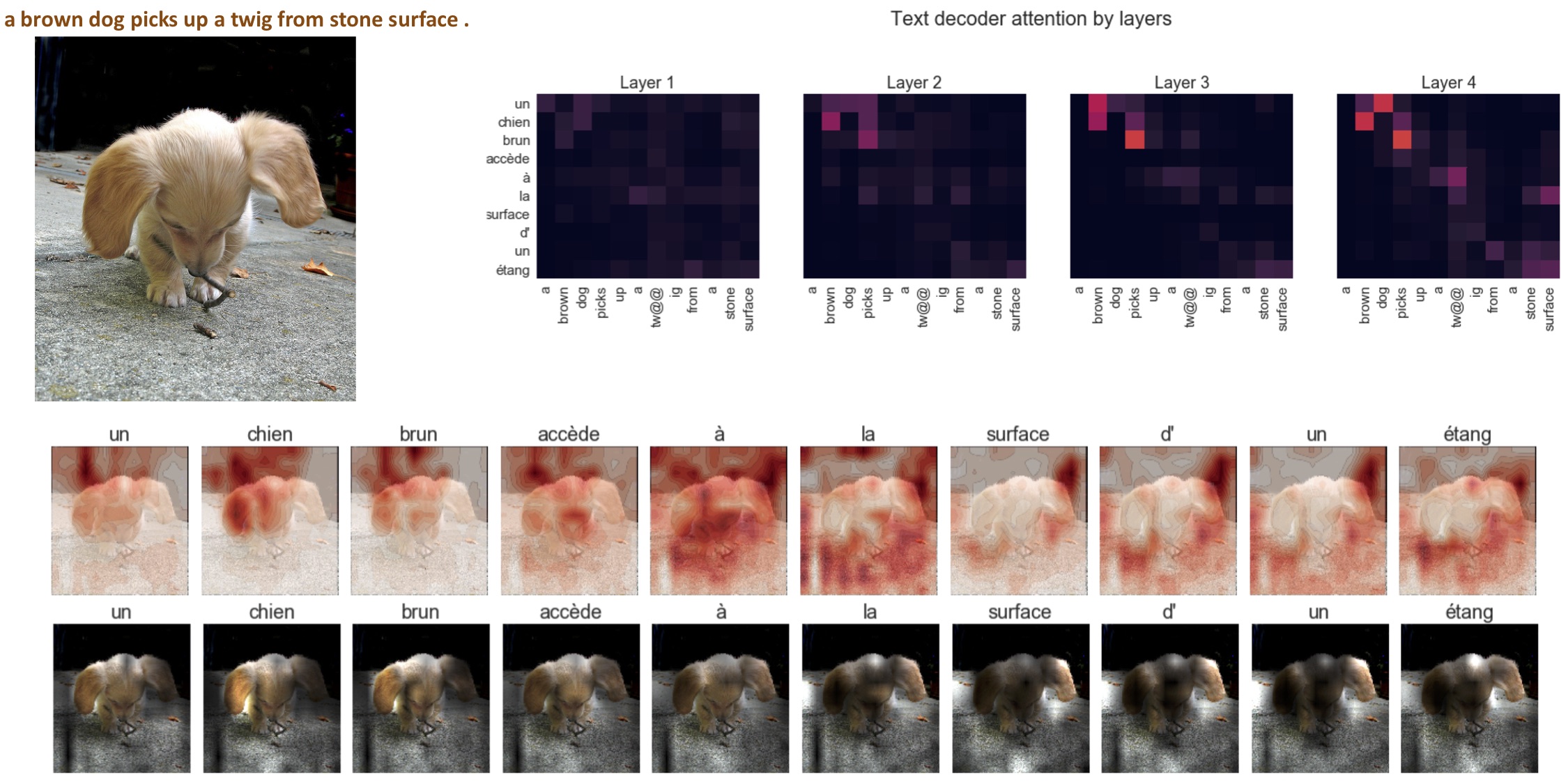}
\caption{Correct attention for \{``chien", ``brun", ``acc\`{e}de" and ``surface"\}, but missed ``twig" for ``\'{e}tang".}
\label{fig:exp_visual_negative}
\end{figure*}

\subsection{Analysis}\label{sec:exp_ana}

In this section, we would shed more light on how and why images can help for unsupervised MT. 
We would first visualize which part of the input image helps the translation by showing the heat map of the transformer attention. 
We then show that image not only helps the translation by providing more information in the testing stage, it can also act as a training regularizer by guiding the model to converge to a better local optimal point in the training stage. 

\subsubsection{Attention}\label{sec:exp_ana_where}

To visualize the transformer's attention from regions in the input image to each word in the translated sentences, we use the scaled dot-production attention of the transformer decoder's multi-head attention block as shown in Figure \ref{fig:model}, more specifically, it is the $\text{softmax}\big( \frac{Q^d_t (K^i)^\text{T}}{\sqrt{d}}\big)$. 
This is a matrix of shape $l_T\times l_S$, where $l_T$ is the translated sentence length and $l_S$ is the source length. Since we flatten the 14$\times$14 matrix from the ResNet152, the $l_S = 196$. 
A heat map for the $j$th word in the translation is then generated by mapping the value of $k$th entry in $\{c_i[j,k]\}^{196}_{k=1}$ to their receptive field in the original image, averaging the value in the overlapping area and then low pass filtering. 
Given this heat map, we would visualize it in two ways: \textbf{(1)} We overlay the contour of the heat-map with the original image as shown in the second, and fifth rows of Figure \ref{fig:exp_visual_positive} and the second row of Figure \ref{fig:exp_visual_negative}; \textbf{(2)} We normalize the heat map between 0 and 1, and then multiply it with each color channel of the input image pixel-wise as shown in the third and sixth rows of Figure \ref{fig:exp_visual_positive} and in the third row of Figure \ref{fig:exp_visual_negative}.
%

We visualize the text attention by simply plotting the text attention matrix $\text{softmax}\big( \frac{Q^d_t (K^e)^\text{T}}{\sqrt{d}}\big)$ in each transformer decoder layer as shown in ``Text decoder attention by layers" in these two figures.

Figure \ref{fig:exp_visual_positive} shows two positive examples that when transformer attends to the right regions of the image like ``orange", ``chapeau", or ``humme" (interestingly, the nose) in the upper image or ``bleu", ``t-shirt", ``blanc" or ``short" in the lower image. 
Whereas in Figure \ref{fig:exp_visual_negative}, transformer attends to the whole image and treat it as a pond instead of focusing on the region where a twig exists. 
As a result, the twig was mistook as pond. 
For the text attention, we can see the text heat map becomes more and more diagonal as the decoder layer goes deeper in both figures. 
This indicates the text attention gets more and more focused since the English and French have similar grammatical rules.

\begin{table}[t]
\small
\centering
\begin{tabular}{l|c|c|c|c}
\hline
Models & En$\rightarrow$  Fr & Fr$\rightarrow$ En & En$\rightarrow$  De & De$\rightarrow$  En  \\ \hline
S-txt  & 6.01 & 6.75 & 6.27 & 6.81 \\
S-txt-img & {\bf 7.55} & {\bf 7.66} & {\bf 7.70} & {\bf 7.53} \\\hline
P-txt & 37.20 & 38.51 & 20.97 & 25.00 \\
P-txt-img & {\bf 39.44} & {\bf 40.30} & {\bf 23.18} & {\bf 25.47} \\\hline
\end{tabular}
\caption{BLEU for testing with \textbf{TEXT ONLY} input}\label{tab:exp_ana_genera}
\end{table}

\begin{table}[t]
\small
\centering
\begin{tabular}{l|c|c|c|c}
\hline
Models & En$\rightarrow$  Fr & Fr$\rightarrow$ En & En$\rightarrow$  De & De$\rightarrow$  En  \\ \hline
S-txt & 13.26  $\uparrow$ & 11.37 $\uparrow$ & 4.15  $\uparrow$& 6.14  $\uparrow$\\
S-txt-img & {\bf 16.10}  $\uparrow$ & {\bf 13.30}  $\uparrow$& {\bf 6.40}  $\uparrow$& {\bf 7.91}  $\uparrow$\\\hline
P-txt & 1.19  $\uparrow$& 1.70  $\uparrow$& 1.39 $\uparrow$ & 2.00 $\uparrow$\\
P-txt-img & {\bf 5.52} $\uparrow$ & {\bf 2.46}  $\uparrow$& {\bf 1.72} $\uparrow$ & {\bf 3.12} $\uparrow$\\\hline
\end{tabular}
\caption{BLEU \textbf{INCREASE} ($\uparrow$) UMNMT model trained on full Multi30k over UMNMT model trained on M30k-half (Table~\ref{tab:exp_stoa} Row 5-8).}\label{tab:exp_ana_reduc}
\end{table}

\subsubsection{Generalizability}\label{sec:exp_ana_genera}

As shown in Equation \ref{eq:mutual}, the model would certainly get more information when image is present in the inferencing stage, but can images be helpful if they are used in the training stage but not readily available during inferencing (which is a very likely scenario in practice)? 
Table \ref{tab:exp_ana_genera} shows that even when images are not used, the performance degradation are not that significant (refer to Row 6-8 in Table \ref{tab:exp_stoa} for comparison) and the trained with image model still outperforms the trained with text only model by quite a margin. 
This suggests that images can serve as additional information in the training process, thus guiding the model to converge to a better local optimal point. 
Such findings also verify the proposed controllable attention mechanism. 
This indicates the requirement of paired image and monolingual text in the testing stage can be relaxed to feeding the text-only data if paired image or images are not available.

\subsubsection{Uncertainty Reduction}\label{sec:exp_ana_reduc}

To show that images help MT by aligning different languages with similar meanings, we also train the UMNMT model on the whole Multi30K dataset where the source and target sentences are pretended unparalleled (i.e., still feed the image text pairs to model). 
By doing this, we greatly increase the sentences in different languages of similar meanings, if images can help align those sentences, then the model should be able to learn better than the model trained with text only. 
We can see from Table \ref{tab:exp_ana_reduc} that the performance increase by using images far outstrip the model trained on text only data, in the case of  En$\rightarrow$ Fr, the P-txt-img has more than 4 points gain than the P-txt. 

\section{Conclusion}

In this work, we proposed a new unsupervised NMT model with multi-modal attention (one for text and one for image) which is trained under an auto-encoding and cycle-consistency paradigm. 
Our experiments showed that images as additional information can significantly and consistently improve the UMT performance. 
This justifies our hypothesis that the utilization of the multi-modal data can increase the mutual information between the source sentences and the translated target sentences.
We have also showed that UMNMT model trained with images can achieve a better local optimal point and can still achieve better performance than trained with text-only model even if images are not available in the testing stage. 
Overall, our work pushes unsupervised machine translation more applicable to the real scenario. 

\newpage

\bibliography{plain.bib}
\bibliographystyle{plain}


\section{Appendices}

\subsection{More Attention Visualization}

\begin{figure*}
\includegraphics[width=\textwidth]{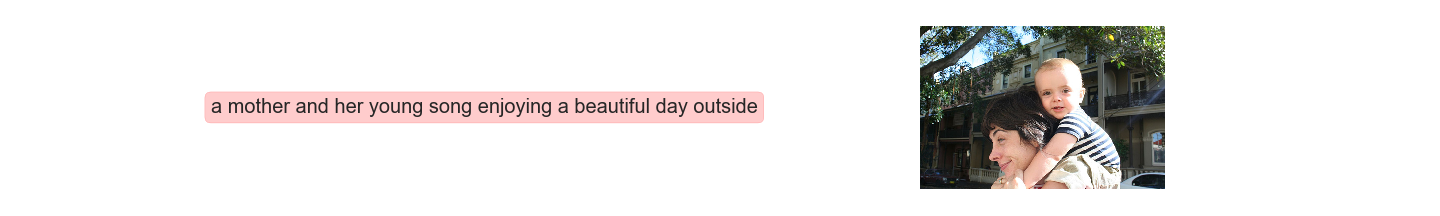}
\includegraphics[width=\textwidth]{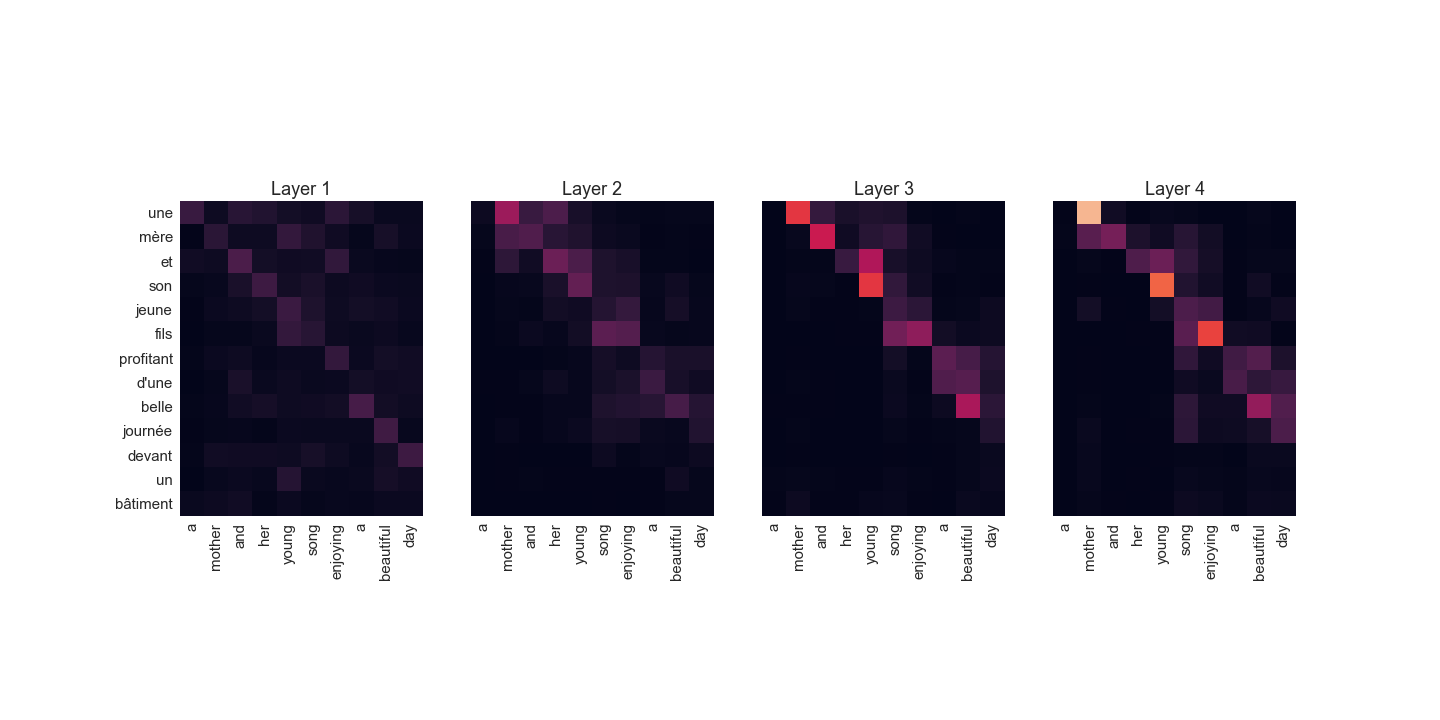}
\includegraphics[width=\textwidth]{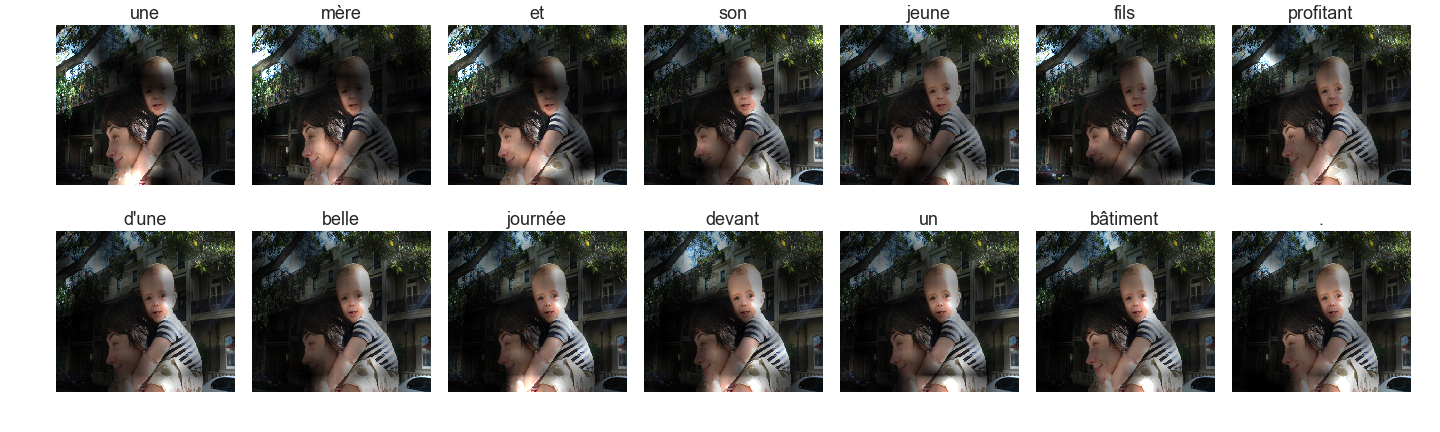}
\includegraphics[width=\textwidth]{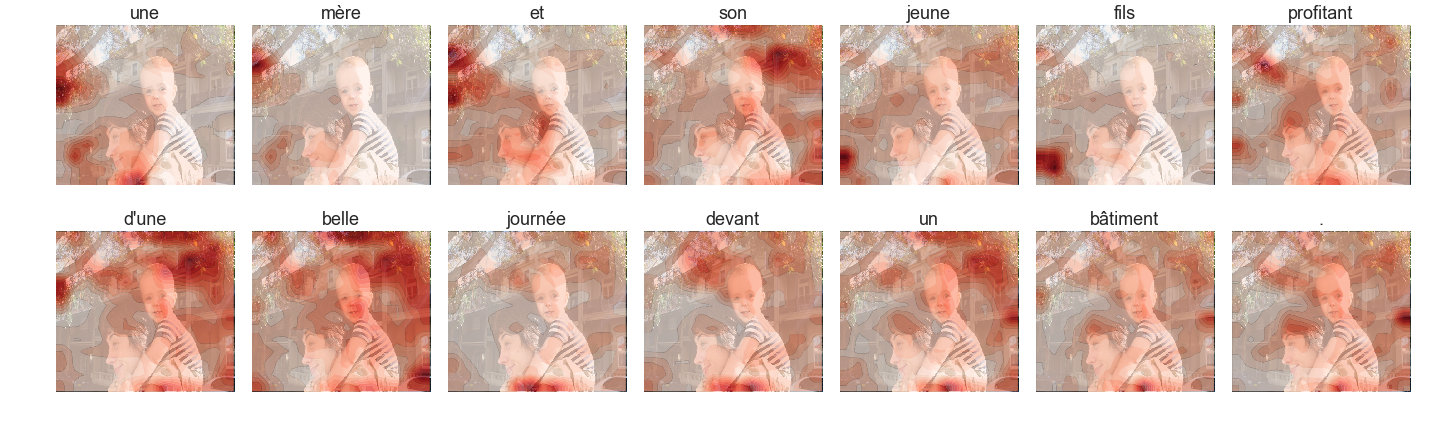}
\end{figure*}

\begin{figure*}
\includegraphics[width=\textwidth]{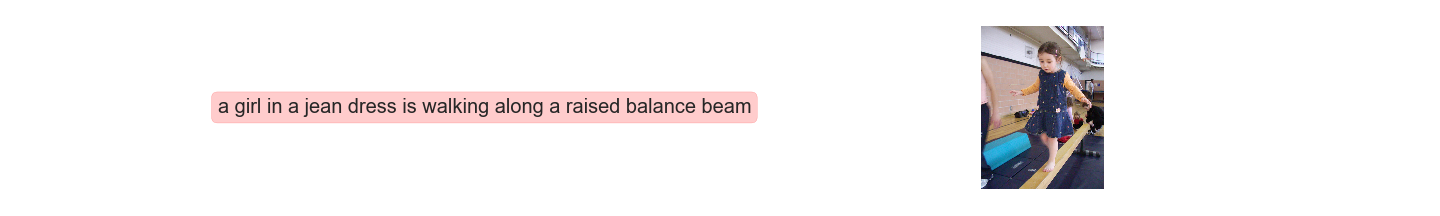}
\includegraphics[width=\textwidth]{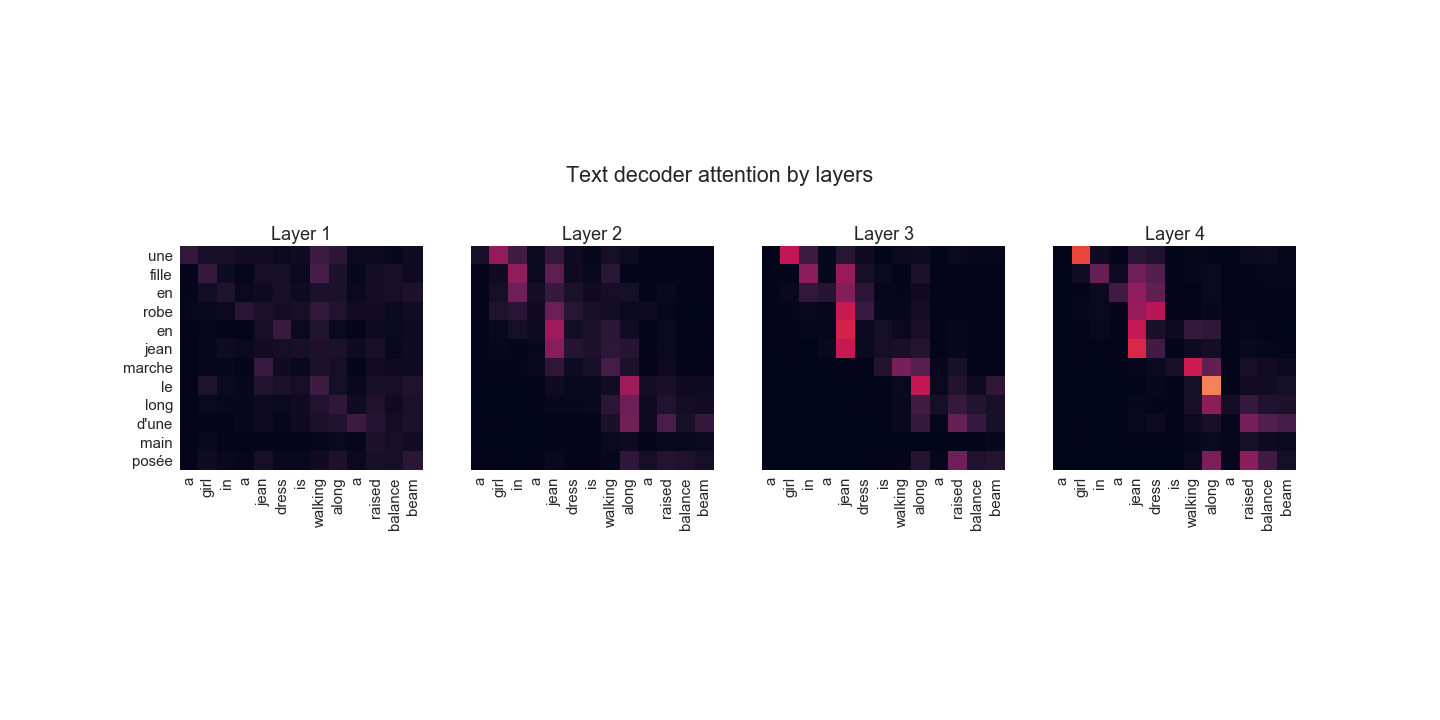}
\includegraphics[width=\textwidth]{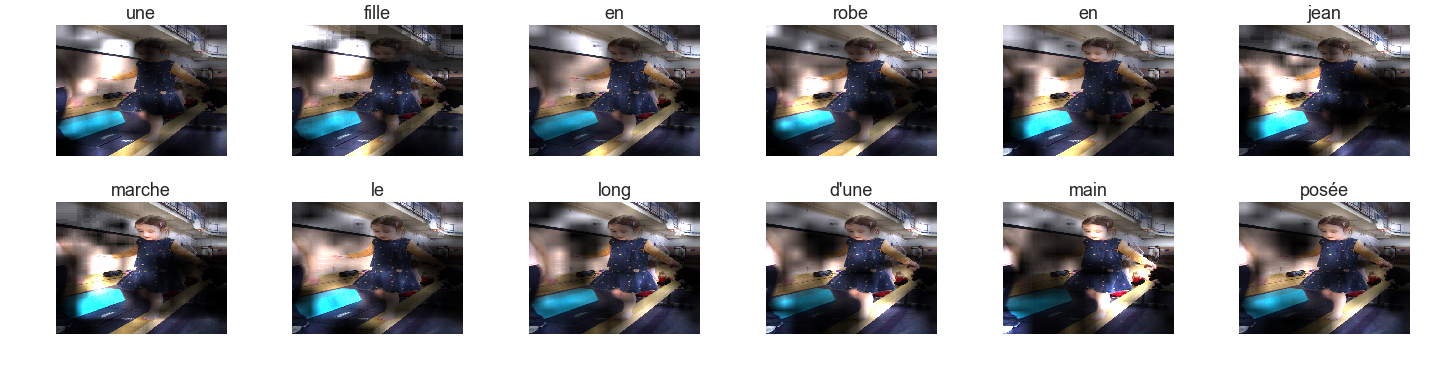}
\includegraphics[width=\textwidth]{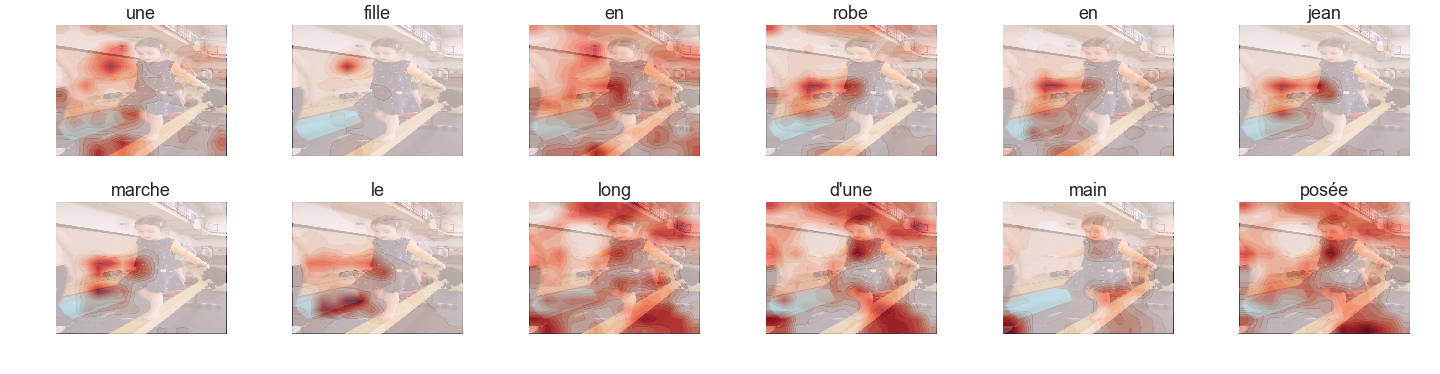}
\end{figure*}

\begin{figure*}
\includegraphics[width=\textwidth]{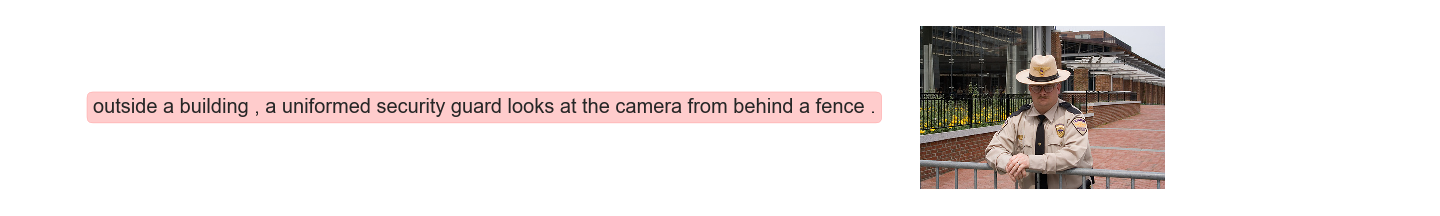}
\includegraphics[width=\textwidth]{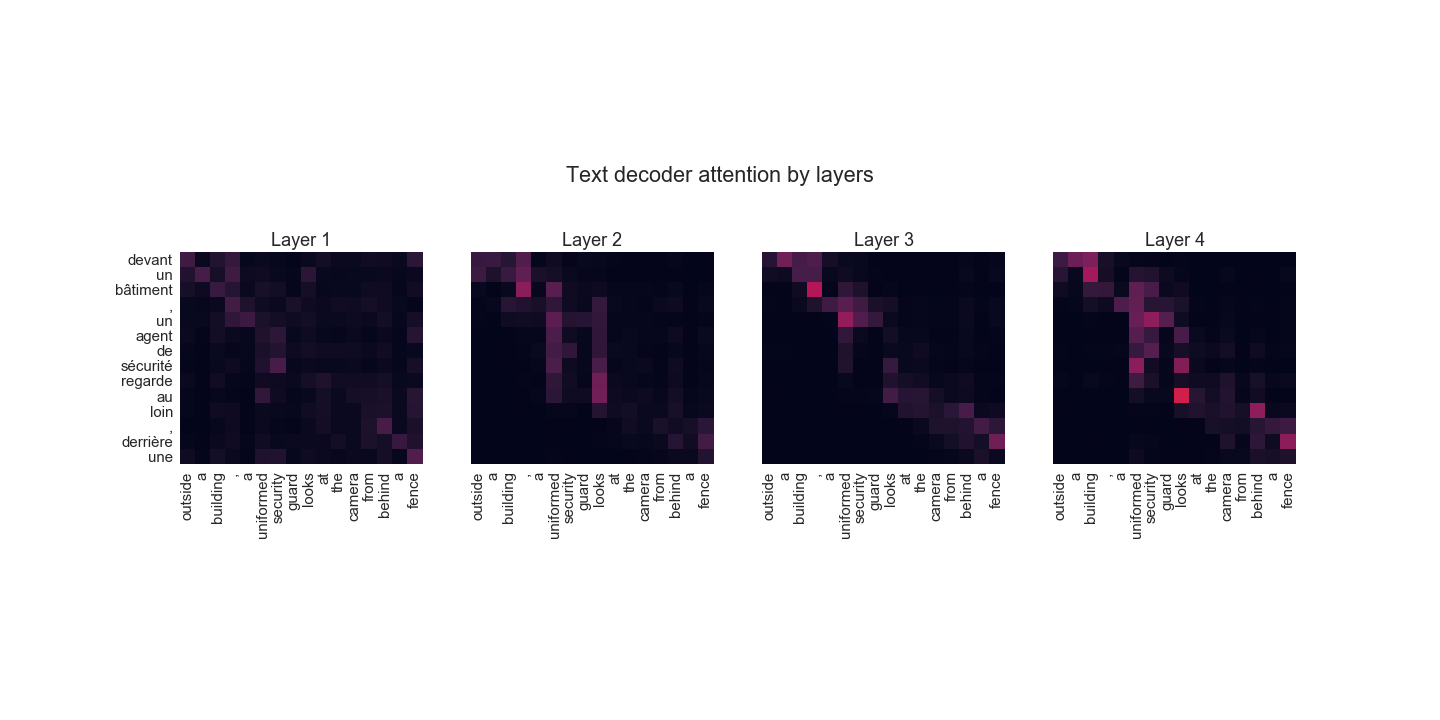}
\includegraphics[width=\textwidth]{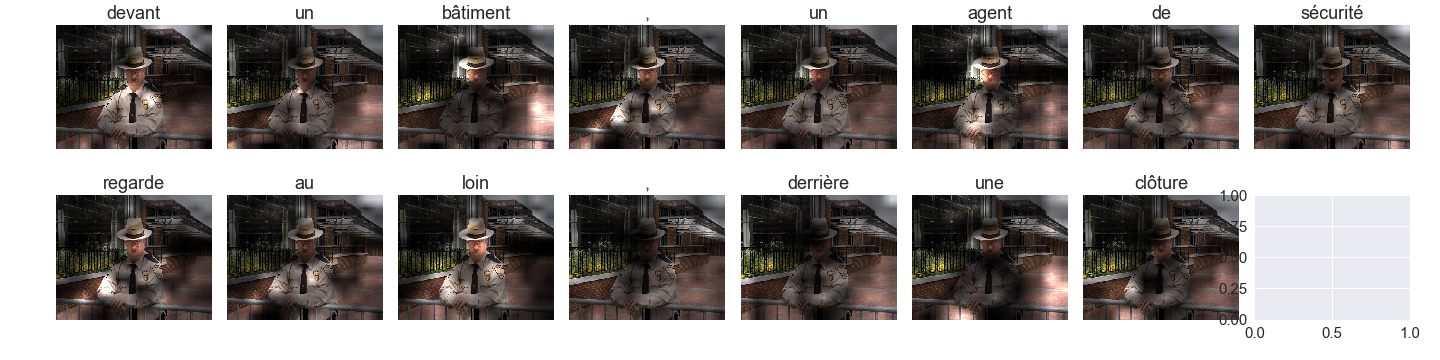}
\includegraphics[width=\textwidth]{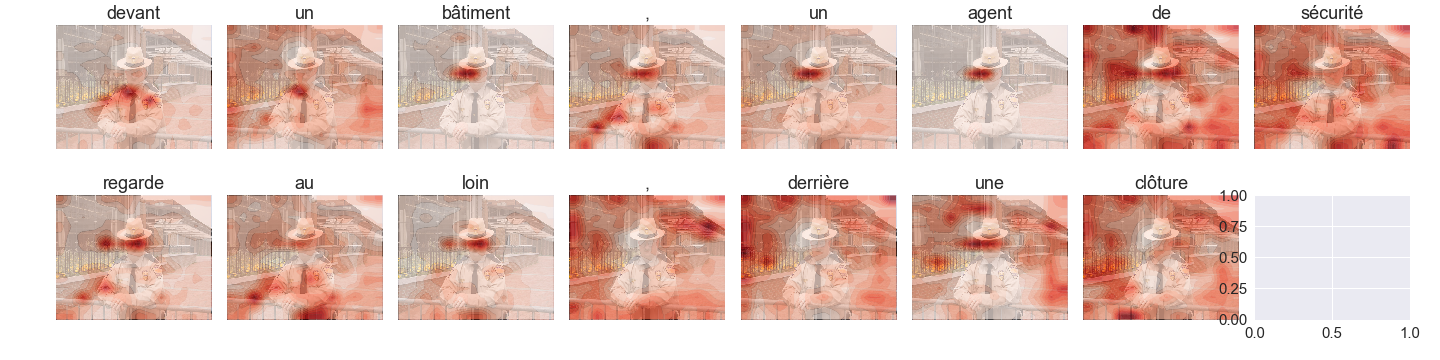}
\end{figure*}

\begin{figure*}
\includegraphics[width=\textwidth]{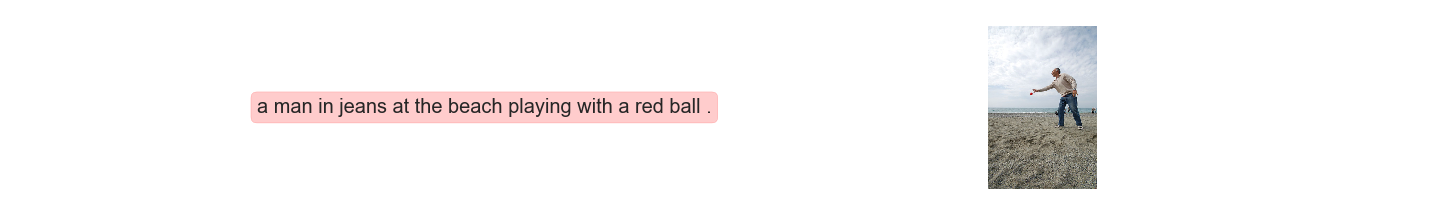}
\includegraphics[width=\textwidth]{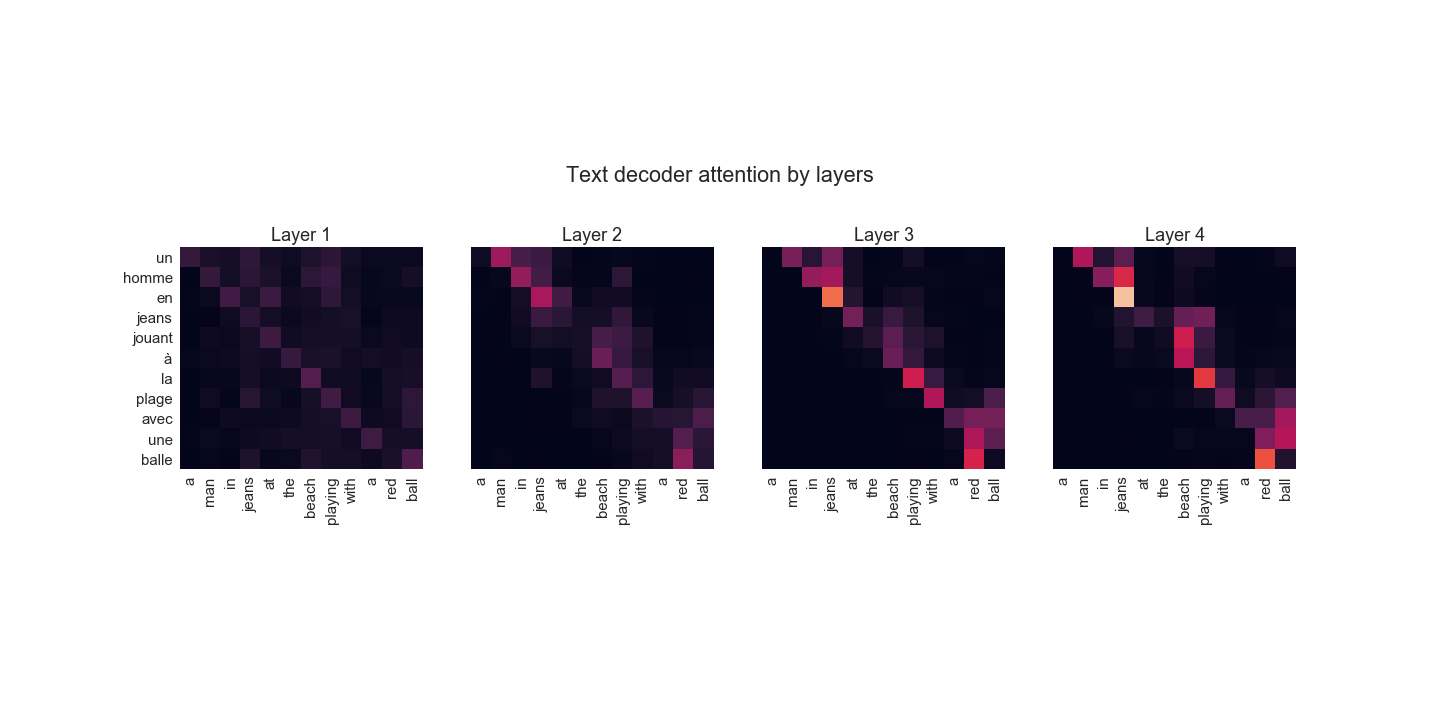}
\includegraphics[width=\textwidth]{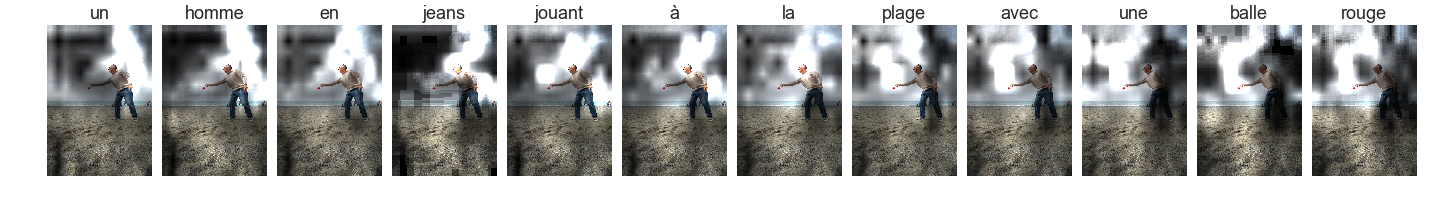}
\includegraphics[width=\textwidth]{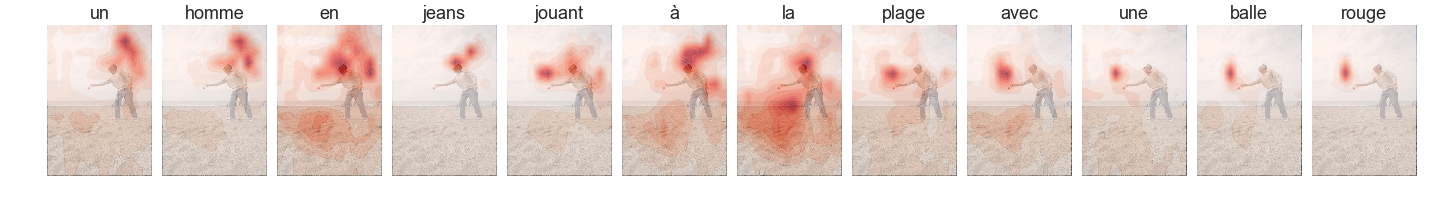}
\end{figure*}

\begin{figure}
\includegraphics[width=\textwidth]{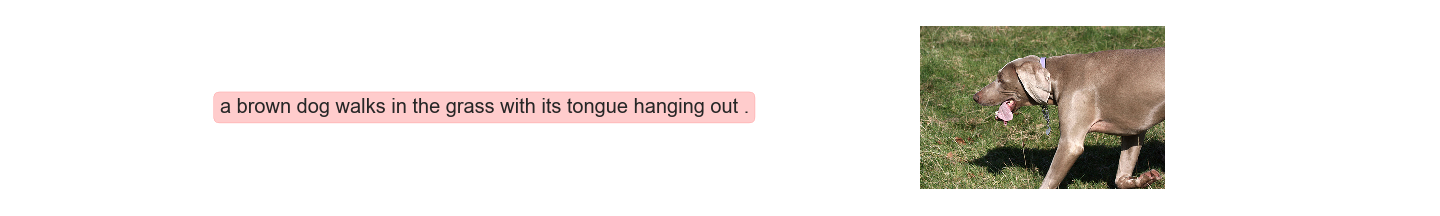}
\includegraphics[width=\textwidth]{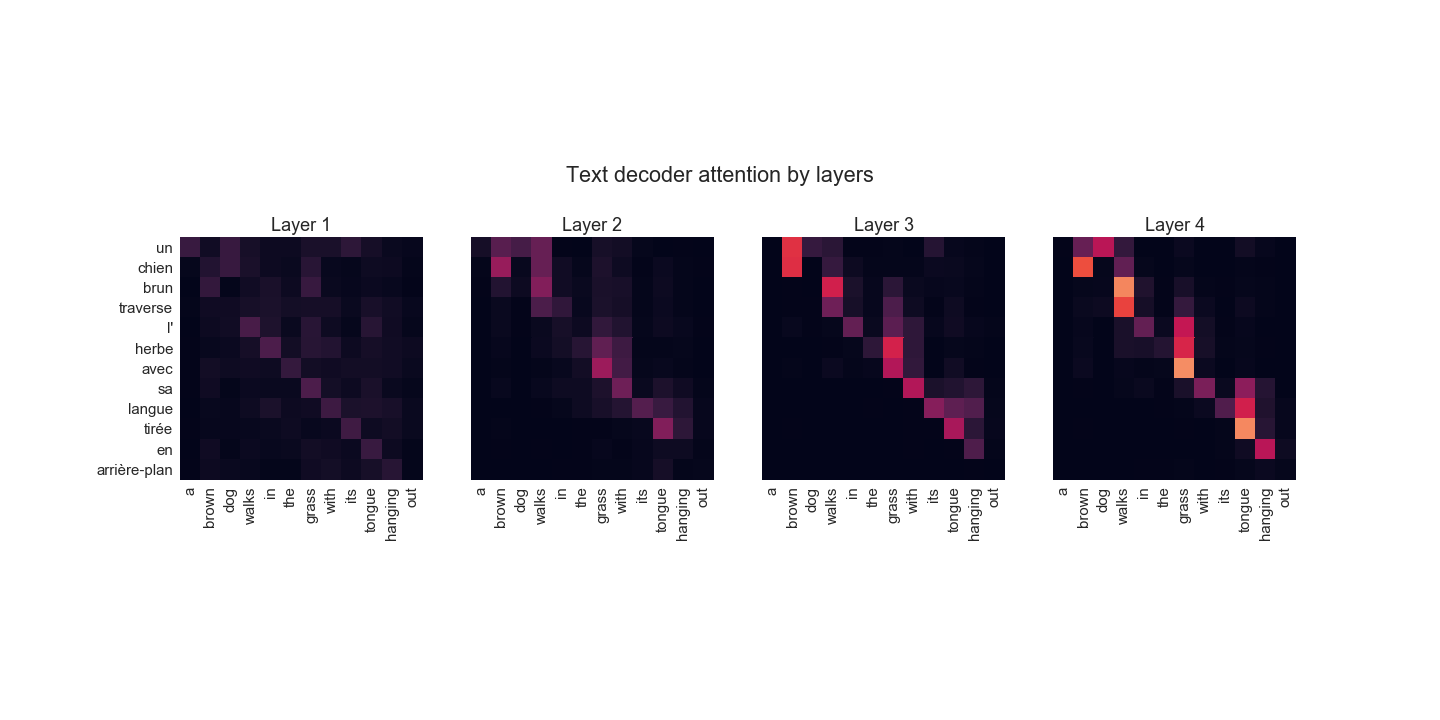}
\includegraphics[width=\textwidth]{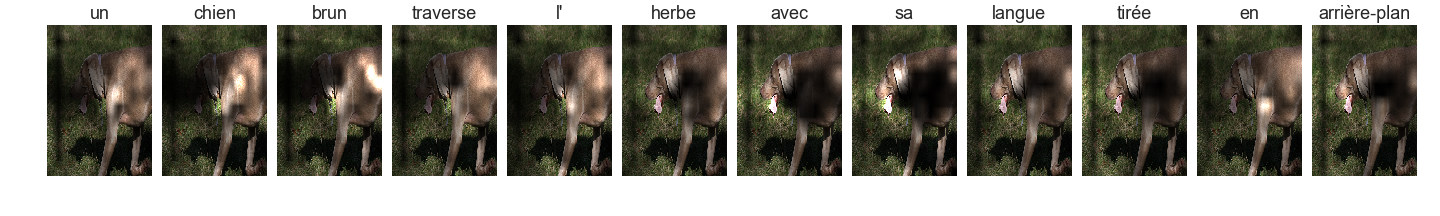}
\includegraphics[width=\textwidth]{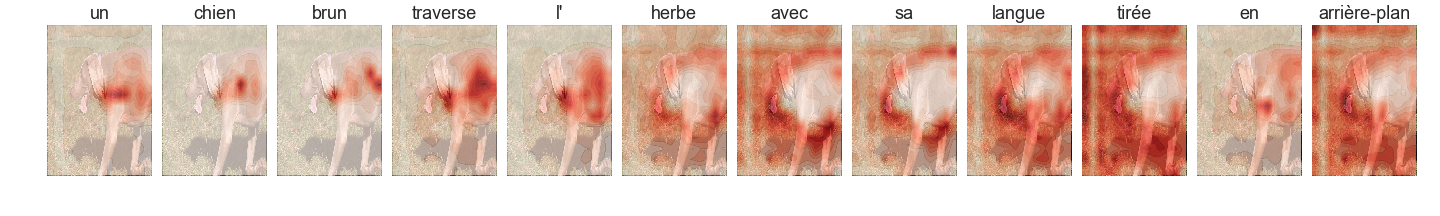}
\end{figure}

\begin{figure*}
\includegraphics[width=\textwidth]{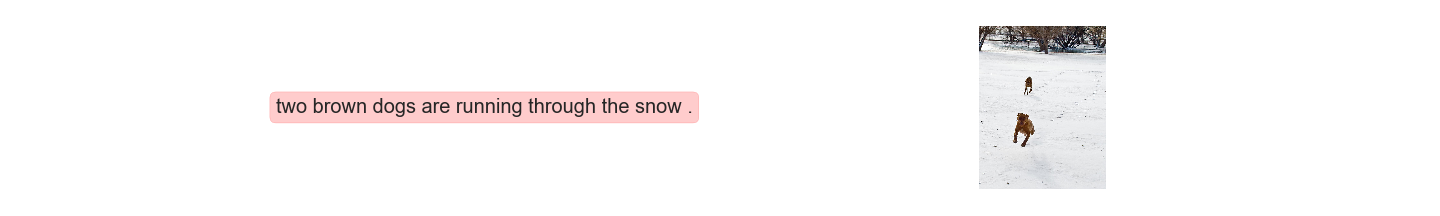}
\includegraphics[width=\textwidth]{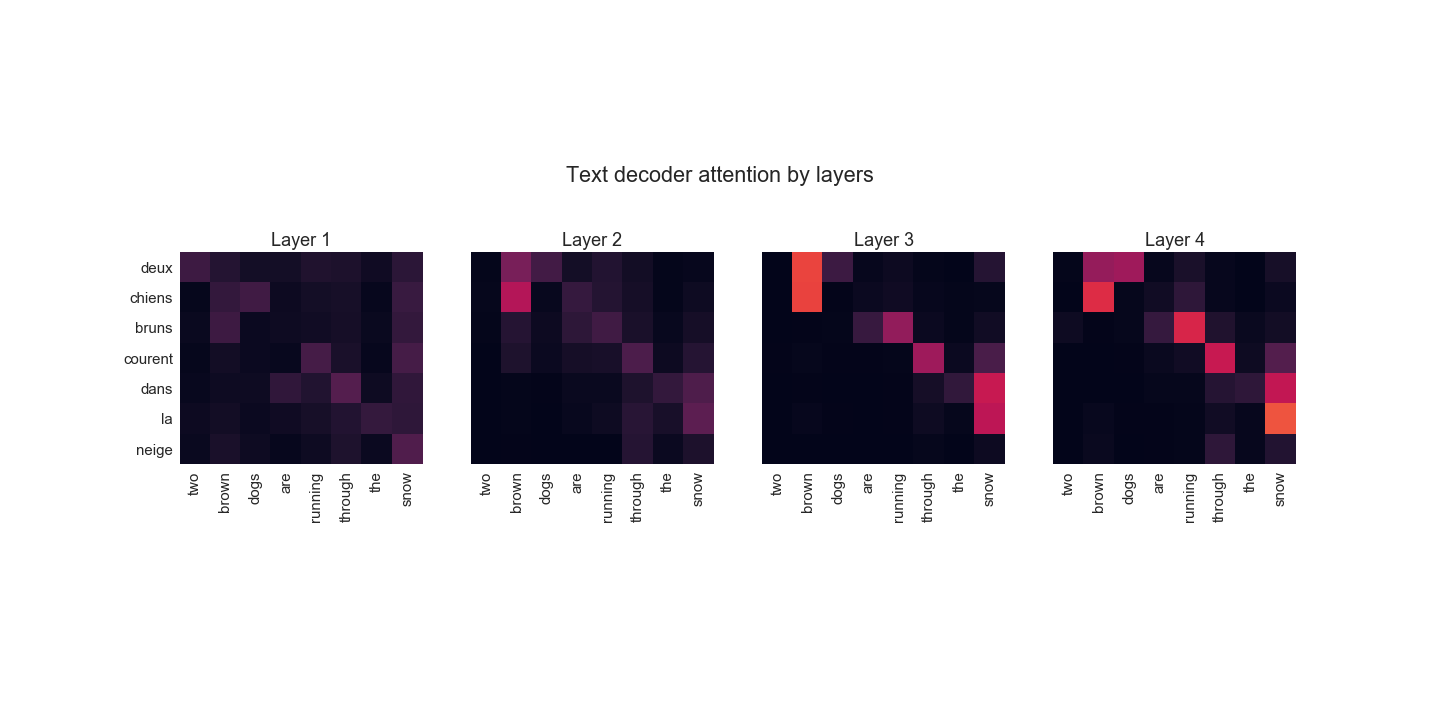}
\includegraphics[width=\textwidth]{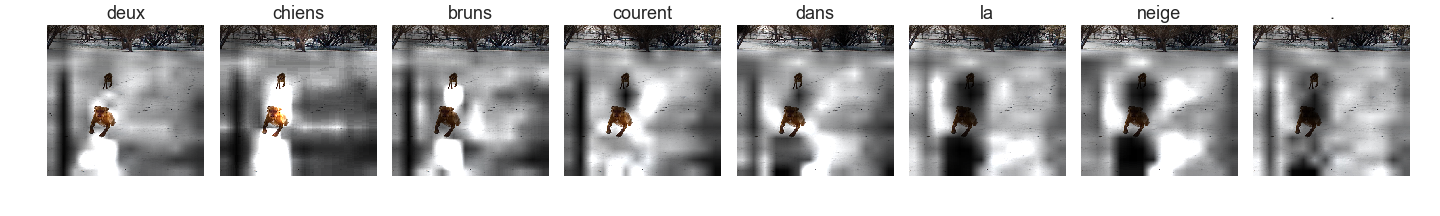}
\includegraphics[width=\textwidth]{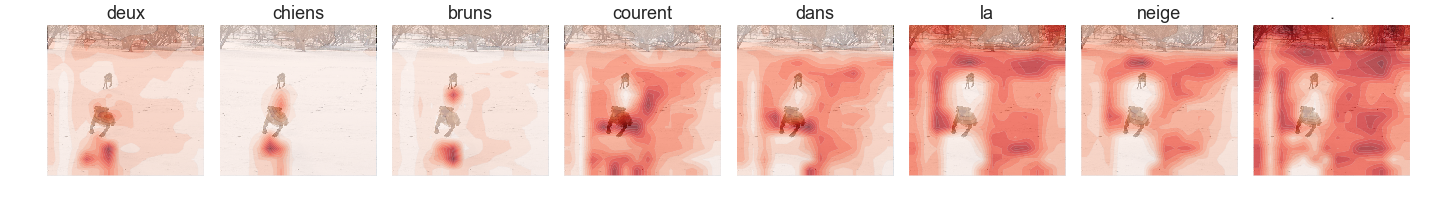}
\end{figure*}

\begin{figure*}
\includegraphics[width=\textwidth]{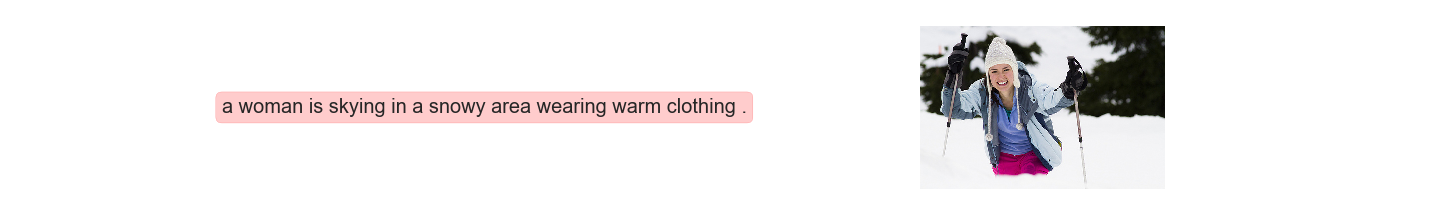}
\includegraphics[width=\textwidth]{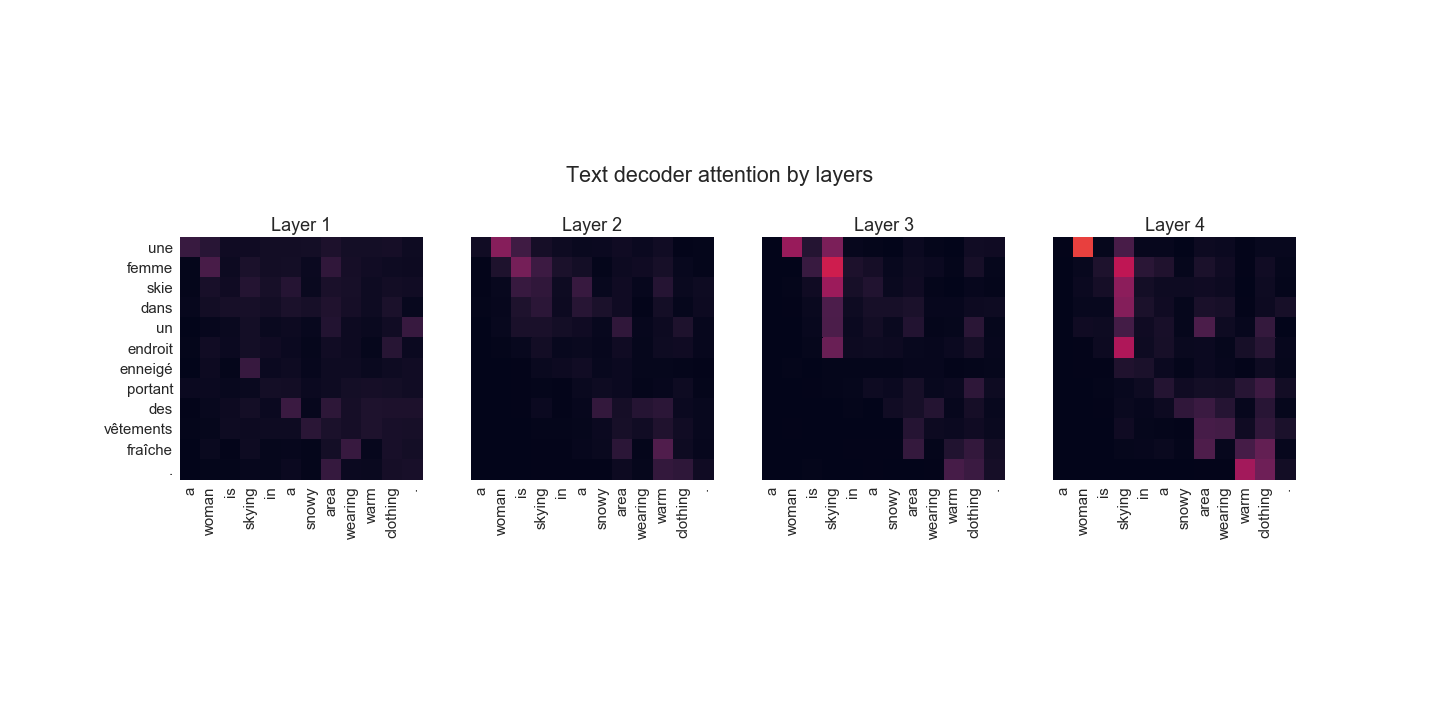}
\includegraphics[width=\textwidth]{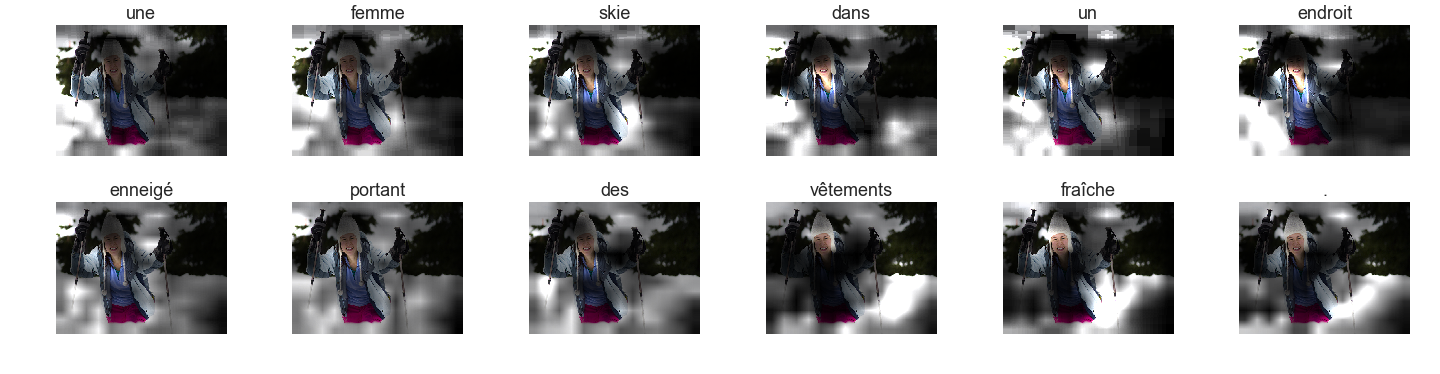}
\includegraphics[width=\textwidth]{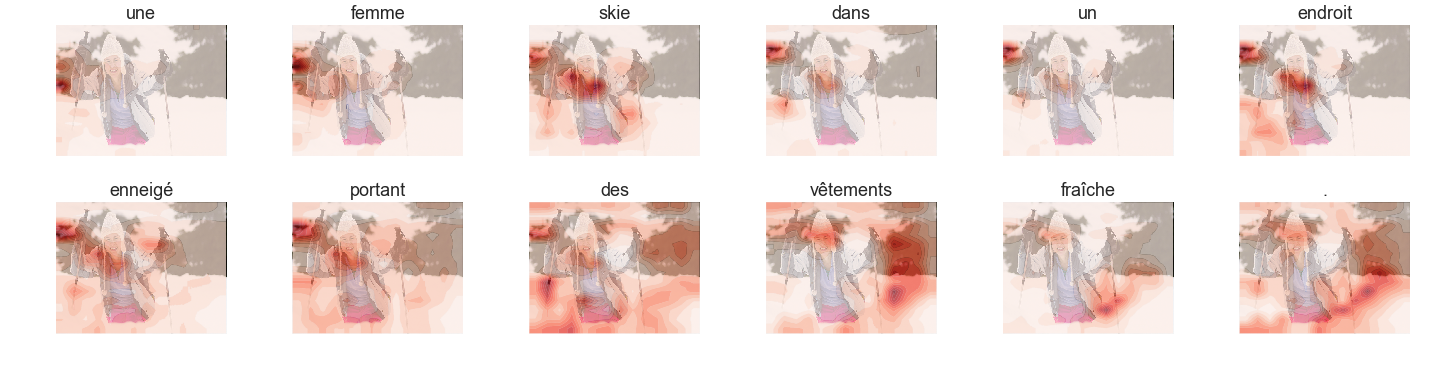}
\end{figure*}

\end{document}